%% file: compiler_paper.tex
\documentclass[journal]{IEEEtran}

\usepackage[english]{babel}

\usepackage{ifpdf}

\usepackage{cite} 
\usepackage{url}
\usepackage{hyperref}
\usepackage{makecell}

\ifCLASSINFOpdf
	\usepackage[pdftex]{graphicx}
	\graphicspath{{./figures/}}
\else
	\usepackage[dvips]{graphicx}
	\graphicspath{./figures/}
\fi
\usepackage{color}
\usepackage{pgf, tikz, pgfplots}
\usepackage{tikz-cd}
\usetikzlibrary{shapes, arrows, automata, plotmarks}
\usetikzlibrary{calc,hobby,decorations}

\usepackage[cmex10]{amsmath}
\usepackage{amsfonts, amssymb, amsthm,bbm}
\usepackage{mathrsfs}

\usepackage{algorithm,algpseudocode}
	\algnewcommand{\LeftComment}[1]{\Statex \(\triangleright\) #1}
\newcommand{\makeboxlabel}[1]{#1\hfill}% \hfill fills the label box

\usepackage{enumerate}
\usepackage{multirow}
\usepackage{rotating}
\usepackage{subcaption}
\input{./latex_resources/mySections.tex}

\input{./latex_resources/mySymbol.sty}
\input{./latex_resources/pennColors.sty}

\def\Hom{\text{Hom}}

\newcolumntype{A}{>{\centering}m{0.12\textwidth}}
\newcolumntype{B}{m{0.09\textwidth}}
\newcolumntype{C}{m{0.085\textwidth}}

\newcolumntype{D}{>{\centering}m{0.06\textwidth}}
\newcolumntype{E}{m{0.06\textwidth}}

\definecolor{mybg1}{rgb}{0.0, 0.6, 0.6}
\definecolor{mybg2}{rgb}{0.0, 0.2, 0.4}

\newtheorem{lemma}{\hspace{0pt}\bf Lemma}

\newtheorem{theorem}{\hspace{0pt}\bf Theorem}

\newtheorem{remark}{\hspace{0pt}\bf Remark}

\newtheorem{method}{\hspace{0pt}\bf Method}

\begin{document}

%%-------------------------------
%%--------- TITLE ---------------
%%-------------------------------
 \title{Graphon Pooling for Reducing Dimensionality of Signals and Convolutional Operators on Graphs}

%%-------------------------------
%%--------- AUTHORS -------------
%%-------------------------------
\author{Alejandro Parada-Mayorga, Zhiyang Wang,
        and Alejandro Ribeiro
\thanks{Preliminary results reported in EUSIPCO2020~\cite{alej2020graphon}. The authors are with the Dept. of Electrical and Systems Eng., University of Pennsylvania. Email: alejopm@seas.upenn.edu, zhiyangw@seas.upenn.edu, aribeiro@seas.upenn.edu.}}

\markboth{IEEE Transactions on Signal Processing (submitted)}%
{Shell \MakeLowercase{\textit{et. al.}}: Bare Demo of IEEEtran.cls for Journals}
\maketitle

%%%%%%%%%%%%%%%%%%%%%%%%%%%%%%%%%%%
%%%%%%%%%%%% ABSTRACT %%%%%%%%%%%%%%%
%%%%%%%%%%%%%%%%%%%%%%%%%%%%%%%%%%%

\input{\pathsections/sec_abstract.tex}


\IEEEpeerreviewmaketitle

%%%%%%%%%%%%%%%%%%%%%%%%%%%%%%%%%%%%%%%%%%%%%%%%%%%%%
%%%%%%%%%%%%%%%%%%%%% SECTIONS  %%%%%%%%%%%%%%%%%%%%%%%%
%%%%%%%%%%%%%%%%%%%%%%%%%%%%%%%%%%%%%%%%%%%%%%%%%%%%%

%%%%%%%%%%%%%%%%%%%%%%%%%%%%%%%%%%
%%%%%%%%% Section: Introduction %%%%%%%%%
%%%%%%%%%%%%%%%%%%%%%%%%%%%%%%%%%%

\input{\pathsections/sec_introduction.tex}

%%%%%%%%%%%%%%%%%%%%%%%%%%%%%%%%%%%
%%%%% Section: Graphs Neural Networks %%%%%%
%%%%%%%%%%%%%%%%%%%%%%%%%%%%%%%%%%%

\input{\pathsections/sec_graphonNN.tex}

%%%%%%%%%%%%%%%%%%%%%%%%%%%%%%%%%%%
%%%%%%%%%% Section: Graphons  %%%%%%%%%%%
%%%%%%%%%%%%%%%%%%%%%%%%%%%%%%%%%%%

\input{\pathsections/sec_graphonSP.tex}

%%%%%%%%%%%%%%%%%%%%%%%%%%%%%%%%%%%
%%%%%%%% Section: Graphon Pooling  %%%%%%%%
%%%%%%%%%%%%%%%%%%%%%%%%%%%%%%%%%%%

\input{\pathsections/sec_graphonPool.tex}

%%%%%%%%%%%%%%%%%%%%%%%%%%%%%%%%%%%
%%5%%%% Section: Numerical Experiments  %%%%%
%%%%%%%%%%%%%%%%%%%%%%%%%%%%%%%%%%%

\input{\pathsections/sec_numsim.tex}

%%%%%%%%%%%%%%%%%%%%%%%%%%%%%%%%%%%
%%%%%%%%%% Section: Discussion  %%%%%%%%%%
%%%%%%%%%%%%%%%%%%%%%%%%%%%%%%%%%%%

\input{\pathsections/sec_discussion.tex}

%%%%%%%%%%%%%%%%%%%%%%%%%%%%%%%%%%%%%
%%%%%% SECTION: APPENDICES  %%%%%%%%%%%%%%
%%%%%%%%%%%%%%%%%%%%%%%%%%%%%%%%%%%%%

\appendices

\input{\pathsections/sec_appendix.tex}

%%%%%%%%%%%%%%%%%%%%%%%%%%%%%%%%%%%%%
%%%%%% SECTION: REFERENCES  %%%%%%%%%%%%%%
%%%%%%%%%%%%%%%%%%%%%%%%%%%%%%%%%%%%%

\bibliographystyle{IEEEbib}
\bibliography{bibliography}

\ifCLASSOPTIONcaptionsoff
  \newpage
\fi

\end{document}

%% file: latex_resources/mySections.tex
\usepackage{needspace}

%% file: v16/sec_abstract.tex
%!TEX root =../compiler_paper.tex

\begin{abstract}

	In this paper we propose a pooling approach for convolutional information processing on graphs relying on the theory of graphons and limits of dense graph sequences. We present three methods that exploit the induced graphon representation of graphs and graph signals on partitions of $[0,1]^2$ in the graphon space. As a result we derive low dimensional representations of the convolutional operators, while a dimensionality reduction of the signals is achieved by simple local interpolation of functions in $L^2 ([0,1])$. We prove that those low dimensional representations constitute a convergent sequence of graphs and graph signals, respectively. The methods proposed and the theoretical guarantees that we provide show that the reduced graphs and signals inherit spectral-structural properties of the original quantities. We evaluate our approach with a set of numerical experiments performed on graph neural networks (GNNs) that rely on graphon pooling. We observe that graphon pooling performs significantly better than other approaches proposed in the literature when dimensionality reduction ratios between layers are large. We also observe that when graphon pooling is used we have, in general, less overfitting and lower computational cost.
	
\end{abstract}

%% file: v16/sec_introduction.tex
%!TEX root =../compiler_paper.tex

%%%%%%%%%%%%%%%%%%%%%%%%%%%%%%%%%%%%%
%%%%%%%%%%%% Section %%%%%%%%%%%%%%%%
%%%%%%%%%%%%%%%%%%%%%%%%%%%%%%%%%%%%%

\section{Introduction}

%-----High level motivation

The problem of pooling entails finding lower dimensional representations of \textit{signals} defined on a given domain and \textit{operators} acting on these signals~\cite{boureau_pooling_1, boureau_pooling_2, goodfellow2016deep, mathDNN_wiatowski, Bronstein2021GeometricDL}. A good pooling technique is such that the result of applying pooled operators to pooled signals yields outputs that are close to pooled versions of the result of applying the original operator to the original signal. In the classical case of signals in time or space processed with convolutional filters or convolutional neural networks (CNNs), pooling is rather straightforward. A pooled signal is a local average and a pooled filter is a sampled filter. Provided that the signal or the filter contain most of their energy in sufficiently low frequencies, the sampling theorem guarantees that processing the pooled signal with the pooled (sampled) filter is a good approximation of the processing of the original signal with the original filter \cite[Section 1.7]{oppenheim1975digital}.

In this paper we consider signals supported on graphs -- large graphs in particular -- and graph convolutional operators in the form of graph filters and graph neural networks (GNNs). In this case, pooling is more involved. To pool signals we can still consider some simple local averages. We can also invoke results from the graph signal processing literature to ascertain that the pooled signal and the original signal are similar representations if the energy of the signal is concentrated in low graph frequencies~\cite{ortega_proxies, tsitsverobarbarossa, bookgspchapsampling, 7055883,7581102, alejopm_bn_j, alejopm_bn_dsw, alejopm_bn_sampta, alejopm_cographs_dsw, alejopm_phd_thesis}. The challenge is that pooled graph filters are not easy to devise. This is because graph convolutions are polynomials on matrix representations of the graph~\cite{ortega2022introduction,gsp_sandryhaila,Pschel2006AlgebraicSP,gsp_sandryhaila_filters,ortega_gsp}. Thus, once we pool information on a subset of nodes it is unclear how to construct a graph linking these nodes to yield graph convolutions that are good approximations of graph convolutions in the original graph ~\cite{gamagnns,Defferrard2016ConvolutionalNN, graphcoarse1}.

%----Previous literature in pooling

To perform pooling on graphs, two main approaches stand out: Multi-scale or multi-level clustering~\cite{Defferrard2016ConvolutionalNN, graphcoarse1} and zero padding~\cite{gamagnns}. In multi-level clustering -- also known as graph coarsening --, families of graphs are derived by grouping subsets of nodes in the original graph. Each cluster is associated to a node in the pooled graph and inter-cluster connectivity determines edges in the pooled graph~\cite{Defferrard2016ConvolutionalNN, graphcoarse1}. Although satisfactory results are obtained with this approach -- tested on GNNs --, the computational cost is high; a fact that hinders applicability to large graphs. With zero padding the dimensionality of signals and operators is reduced by zeroing specific components of the signal while retaining the original graph~\cite{gamagnns}. This procedure forces a reduction in the dimension of the signal  while inducing a reduction of the effective dimension of the filtering operators. The effectiveness of zero padding depends on the effectiveness of the choice of the set of zeroed nodes. Finding good sets of nodes to zero is computationally expensive and, as is the case of graph coarsening, precludes application to large graphs.

%----- Our contributions ------

In this paper we consider graphs with large numbers of nodes. We leverage the concept of graphons as graph limits and build on the theory of graphon signal processing~\cite{Diao2016ModelfreeCO,Morency2020GraphonFG,ruiz2020graphon} to provide the following contribution:

\smallskip

\begin{list}
      {}
      {\setlength{\labelwidth}{22pt}
       \setlength{\labelsep}{0pt}
       \setlength{\itemsep}{0pt}
       \setlength{\leftmargin}{22pt}
       \setlength{\rightmargin}{0pt}
       \setlength{\itemindent}{0pt} 
       \let\makelabel=\makeboxlabel
       }

\item[{\bf (C1)}] We propose low computational cost pooling methods for signals and operators on graphs based on graphon representations (Section~\ref{sec_graphon_pooling}).

\end{list}

\smallskip\noindent In particular, we derive an operation of pooling by building sequences of graphs and graph signals that converge to a graphon and a graphon signal, respectively. To build these sequences we leverage results about partitions in graphon spaces, performing three simple operations: (i) integration on a regular grid, (ii) integration on an irregular grid, and (iii) random sampling. The graphs obtained by these methods define the reduced operators on the graph, while the reduction of the signal is achieved by simple local interpolation of functions on $L^{2}([0,1])$. 

However simple, graphon pooling methods can yield good pooled representations. We prove that this is true by providing the following two contributions:

\smallskip

\begin{list}
      {}
      {\setlength{\labelwidth}{22pt}
       \setlength{\labelsep}{0pt}
       \setlength{\itemsep}{0pt}
       \setlength{\leftmargin}{22pt}
       \setlength{\rightmargin}{0pt}
       \setlength{\itemindent}{0pt} 
       \let\makelabel=\makeboxlabel
       }

\item[{\bf (C2)}] We consider pooled versions of signals and filters and provide an error bound for the difference between the outcome of processing the pooled signal with the pooled filter and the outcome of processing the original signal with the original filter. These bounds require conditions on graphon filters that are akin to the low frequency conditions that appear in the processing of time signals (Theorems~\ref{thm_change_filters} and \ref{theorem:stabilityAlgNN0}). 

\smallskip

\item[{\bf (C3)}] We show that the shift operators obtained by means of graphon pooling based on integration on a regular grid are stable with respect to arbitrary approximations of the graphon (Theorem~\ref{thm_estimate_graphons}).

\end{list}

\smallskip\noindent Contribution (C3) is important because contribution (C2) requires access to the graphon from which graphs are sampled. Indeed, our results show that graphon pooling methods induce filter perturbations that are bounded by the cut-norm distance between the original graph and its pooled version provided that both are sampled from the same graphon. In practice, graphons must be estimated from graphs. Contribution (C3) implies that graphon estimation errors stay contained when they are mapped to filter perturbation bounds of pooled operators. 

The performance of graphon pooling is tested on GNNs, allowing a direct comparison with the graph pooling methods proposed in~\cite{gamagnns, Defferrard2016ConvolutionalNN, graphcoarse1}. A set of numerical experiments corroborate our findings and show that when the dimensionality reduction of the signals and operators is large, graphon pooling leads to better performance than other approaches (Section~\ref{sec_numsim}). In particular, the best performance is obtained by the graphon pooling method based on integration on a regular grid. This is partly explained by the stability of that method to approximations of the graphon. 

%-----Organization of the paper

This paper is organized as follows. In Section~\ref{sec_graphconv_gnns} we discuss the basics of graph signal processing and GNNs, including the concepts of signals, convolutional operators (filters), and GNN mapping operators. This will provide the scenario where graphon pooling is naturally applied and where it is also going to be tested numerically. Section~\ref{sec_graphons} contains the basics about graphons, graph limits, graphon neural networks (Gphon-NN), and their connection to graphs and GNNs. In Section~\ref{sec_graphon_pooling} we introduce the graphon pooling methods proposed in this paper, while in Section~\ref{sec_numsim} we provide numerical simulations to evaluate the performance of graphon pooling against other pooling approaches. In Section~\ref{sec_discussion} we present discussions and conclusions.

%% file: v16/sec_graphonNN.tex
%!TEX root =../compiler_paper.tex

%%%%%%%%%%%%%%%%%%%%%%%%%%%%%%%%%%%%%%%%%%%%%%
%%%%%%%%%%%%%%%% SECTION %%%%%%%%%%%%%%%%%%%%%
%%%%%%%%%%%%%%%%%%%%%%%%%%%%%%%%%%%%%%%%%%%%%%

\section{Graph Signals, Convolutional Operators on Graphs, and GNNs}\label{sec_graphconv_gnns}

In this section we provide a basic description of signals and convolutional operators on graphs. Additionally, we discuss GNNs and their mapping operators.

%%%%%%%%%%%%%%%%%%%%%%%%%%%%%%%
%%%%%%%%% Sub-Section %%%%%%%%%%%%
%%%%%%%%%%%%%%%%%%%%%%%%%%%%%%%

\subsection{Graph signal processing}

Let us consider the graph $G=(V(G),E(G),w_G)$ with set of vertices $V(G)$, set of edges $E(G)$, and weight function $w_G: E(G)\rightarrow\mathbb{R}^{+}$. We define a graph signal $\mathbf{x}$ on $G$ as the map $\mathbf{x}:V(G)\rightarrow\mathbb{R}$ identified with a vector in $\mathbb{R}^{\vert V(G)\vert}$. The $i$-th component of $\bbx$ is the value of $\bbx$ on the $i$-th node in $V(G)$, given an ordering of $V(G)$. In what follows we will use the symbol $(G,\bbx)$ to denote the graph signal $\bbx$ defined on $G$. We recall that the image set of $E(G)$ under $w_G$ can be stored in a weight or adjacency matrix $\mathbf{A}$, with $\mathbf{A}(i,j)=w_G (\{i,j\})$, $\{i,j\}\in E(G)$. 

In graphs, the notion of filtering and convolution with filters relies on a shift operator $\mathbf{S}\in\mathbb{R}^{N\times N}$ with $N=\vert V(G)\vert$, which can be selected as the Laplacian matrix, normalized Laplacian, or the adjacency matrix $\mathbf{A}$~\cite{gsp_sandryhaila,gsp_sandryhaila_filters,ortega_gsp}. In this paper we consider $\mathbf{S}=\mathbf{A}$ as the shift operator. Then, filters on the graph are polynomial matrix operators given by
\begin{equation}
\mathbf{H}(\mathbf{S})
     =
      \sum_{k=0}^{K-1}h_{k}\mathbf{S}^{k}  
      ,
\end{equation}
where the coefficients $h_{k}$ are called the filter taps. The convolution between a filter $\mathbf{H}(\mathbf{S})$ and a signal $(G,\mathbf{x})$ results in a signal $(G,\bby)$ with
\begin{equation}\label{eq:Hingraph}
\bby =
\mathbf{H}(\mathbf{S})\mathbf{x}
          =
          \sum_{k=0}^{K-1}h_{k}\mathbf{S}^{k}\mathbf{x}
          .
\end{equation}
%
%
%

%%%%%%%%%%%%%%%%%%%%%%%%%%%%%%%%%%%%%
%%%%%%%%%%%% Sub-Section %%%%%%%%%%%%
%%%%%%%%%%%%%%%%%%%%%%%%%%%%%%%%%%%%%

\subsection{Graph neural networks}

%%-----------------------------
%%--------- FIGURE ------------
%%-----------------------------

\begin{figure}
\centering
\includegraphics[width=0.5\textwidth]{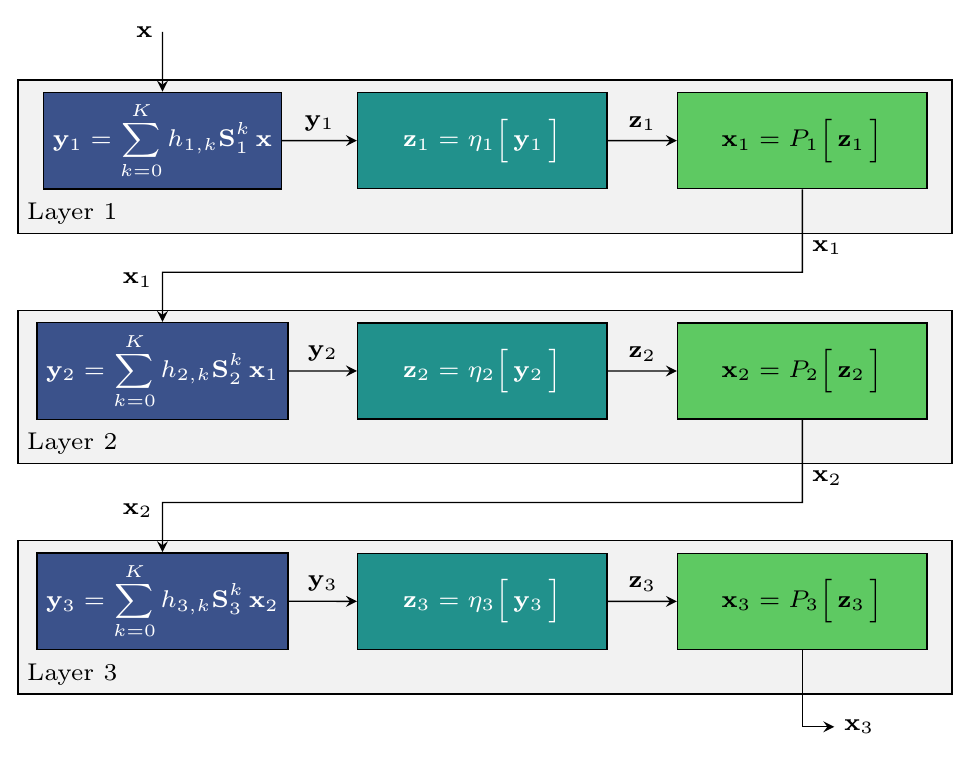}
\caption{Graph neural network with three layers. The input signal $\mathbf{x}$ is processed by the GNN to produce an output $\bbx_3$. In the $i$-th layer the information is processed by a convolution operator $\sum_{k=0}^{K}h_{i,k}\bbS_{i}^k$, then by a pointwise nonlinearity $\eta_i$, and then by a pooling operator $P_i$ that matches the dimensions of the signals between layers.}
\label{fig:GNNbasicfig}
\end{figure}

%%-------- End of Figure ---------

A graph neural network (GNN) is a stacked layered structure -- see Fig.~\ref{fig:GNNbasicfig}. In each layer, information is processed by means of convolutional operators acting on graph signals, followed by pointwise nonlinearities and pooling operators. Then, in the $i$-th layer of the GNN an input signal $\bbx_{i-1}$ is filtered as $\bby_i = \sum_{k=0}^{K}h_{i,k}\bbS_{i}^{k}\bbx_{i-1}$. Afterwards, a point-wise nonlinearity operator $\eta_i$ is applied to $\bby_i$ to obtain $\bbz_i = \eta_{i}(\bby_i)$. Finally, a pooling operator $P_i$ reduces the dimension of $\bbz_i$ and the size of $G_i$ -- which implies changes in $\bbS_i$ -- generating $\bbx_i = P_i (\bbz_{i})$. The operation defined by $P_i$ is meant to preserve structural properties of information, and the function $\eta_{i}$ is required to be Lipschitz~\cite{fern2019stability,gamagnns,mathDNN_wiatowski}. The output of the $i$-th layer can be written as
%
%
%
%\begin{equation}\label{eq:Hingnn}
$
\mathbf{x}_{i}
           =
              P_i \left(  
                              \eta_i \left( 
                                                        \bbH_i (\bbS_i) \bbx_{i-1}
                                              \right)
                    \right)
                    ,
                    ~
                    \text{with}
                    ~
                    \bbH_i (\bbS_i) = \sum_{k=0}^{K}h_{i,k}\bbS_{i}^{k}
                    .
$                    
%\end{equation}
%
%
%
In each layer several filters can be considered to produce outputs of multiple features. 

Since the pooling operator $P_i$ reduces the dimensionality of the data, this entails a possible modification of the underlying graphs in the layers. Among the pooling methods considered for GNNs two approaches stand out, graph coarsening and zero padding. When graph coarsening is used, spectral clustering techniques are used to group subsets of nodes and edges in a graph $G_i$ to define a new graph $G_{i+1}$ where $\vert V(G_{i+1})\vert < \vert V(G_{i})\vert $~\cite{Defferrard2016ConvolutionalNN,graphcoarse1}. To do graph pooling with zero padding, the dimensions of the graphs are preserved while some components of the signals are forced to be zero~\cite{gamagnns}. To optimally select those components, we use graph sampling approaches like those proposed in~\cite{ortega_proxies,tsitsverobarbarossa,bookgspchapsampling,7055883,7581102,alejopm_bn_j,alejopm_bn_dsw,alejopm_bn_sampta,alejopm_cographs_dsw,alejopm_phd_thesis}. Then, while zero padding does not modify the dimensions of the graphs, i.e. $\vert V(G_{\ell+1})\vert = \vert V(G_{\ell})\vert $, it does implicitly modify the shift operators, which are now associated to an \textit{induced subgraph} of the original graph.

The coefficients of the convolutional operators $h_k$ are learned from the data. This is, for a training set $\mathcal{T}=\{(\mathbf{x},\mathbf{y})\}$ with inputs $\mathbf{x}$ and outputs $\mathbf{y}$, the GNN learns a representation mapping that associates an output $\hat{\mathbf{y}}$ for a given input $\hat{\mathbf{x}}$ with $(\hat{\mathbf{x}},\hat{\mathbf{y}})\notin\mathcal{T}$~\cite{gamagnns}.

To represent the mapping operator of a GNN with $L$ layers we use the symbol $\mathbf{\Phi}\left( \bbx, \{ \ccalF_\ell \}_{\ell=1}^{L}, \{ \bbS_\ell \}_{\ell=1}^{L}\right)$, where $\bbx$ is the input signal to the GNN, $\ccalF_\ell$ is the set of filters used in the $\ell$-th layer, and $\bbS_\ell$ is the $\ell$-th shift operator.

% %%---------------------------------
% %%------------ Remark --------------
% %%---------------------------------

% \begin{remark}\label{rmk_pooling}\normalfont

% It is worth pointing out that when considering convolutional neural networks whose layers are defined in Euclidean domains, pooling is reduced to the application of sampling operations that are well defined and are associated to the classical tools. In contrast, when considering convolutional architectures whose domains are non-Euclidean the notion of sampling or reduction of the domain is far from trivial~\cite{Bronstein2021GeometricDL}. Indeed, when we consider signals defined on the graph-like domains, the main challenges reside in that these domains are redefined once the dimensionality of the signal is reduced. 

% \end{remark}

% %%--------- End of Remark ------------

%% file: v16/sec_graphonSP.tex
%!TEX root =../compiler_paper.tex

%%%%%%%%%%%%%%%%%%%%%%%%%%%%%%%%%%%%%
%%%%%%%%%%% SECTION %%%%%%%%%%%%%%%%%%%
%%%%%%%%%%%%%%%%%%%%%%%%%%%%%%%%%%%%%

\section{Signals and Convolutional Operators on Graphons}\label{sec_graphons}

A graphon is a symmetric bounded function $W:[0,1]^{2}\rightarrow [0,1]$ that can be conceived as a \textit{completion} of the limits for infinite sequences of graphs in the same way that irrational numbers complete the real line~\cite{Diao2016ModelfreeCO,lovaz2012large,Glasscock2015WhatIA}. Some graphons can be obtained from labeled graphs. For instance, given a graph $G$ with weight matrix $\bbA$ one can obtain an \textit{induced} graphon $W_G$ given by
\begin{equation}
W_{G}(x,y) 
           =
              \mathbf{A}\left( 
                                              \lceil Nx\rceil, \lceil Ny\rceil
                                     \right)
                                     ,                                    
\end{equation}
where $\lceil \cdot \rceil$ is the ceiling operator, $N = \vert V(G) \vert$, and $x,y \in [0,1]$. Therefore, $W_G$ is a piece-wise constant function on a regular grid, and the amplitudes of $W_G$ are given by the weight matrix $\bbA$.

%%%%%%%%%%%%%%%%%%%%%%%%%%%%%%%%%%
%%%%%%%%%%% Sub-Section %%%%%%%%%%%%%%
%%%%%%%%%%%%%%%%%%%%%%%%%%%%%%%%%%

\subsection{Convergence of graphs to graphons}

The concept of convergence of sequences of graphs to graphons relies on the convergence of sequences of homomorphism densities -- see subsection~\ref{subsec_basic_preliminaries} in the Appendix. This type of convergence is compatible with a notion of convergence based on the cut norm $\Vert \cdot \Vert_{\square}$. To see this, let us consider the metric distance $\delta_{\square}(W_{G_i}, W)$ between $W_{G_i}$ and $W$ given by
%
%
%
%\begin{equation}
$
\delta_{\square}(W_{G_i}, W) 
           = 
             \inf_{\pi}
                      \left\Vert 
                                    W
                                     -
                                   \pi(W_{G_i})
                       \right\Vert_{\square} 
,
$
%\end{equation}
%
%
%
where $\pi: [0,1]\rightarrow [0,1]$ is a measure preserving map -- analogous to a node label permutation on the graph --, and the cut norm is given by
\begin{equation}\label{eq_cutnorm_def}
\left\Vert 
W(x,y)
\right\Vert_{\square}
=
\sup_{I,J\subset [0,1] } 
\left\vert 
\int_{I\times J} W(x,y) dxdy
\right\vert
.
\end{equation}
As shown in~\cite{lovaz2012large,Diao2016ModelfreeCO}, if $\{ G_i \}_{i}$ converges to $W(x,y)$ in the homomorphism sense -- which we denote by  $\{ G_i \}\rightarrow W(x,y)$ --, then $\delta_{\square}(W_{G_i},W)\rightarrow 0$. Notice that the action of $\pi$ on $W_{G_i}(x,y)$ is given by $\pi (W_{G_i}(x,y)) = W_{G_i}(\pi(x),\pi(y))$.

%%%%%%%%%%%%%%%%%%%%%%%%%%%%%%%%%%%%%%%%%%
%%%%%%%%%%% Sub-Section %%%%%%%%%%%%%%
%%%%%%%%%%%%%%%%%%%%%%%%%%%%%%%%%%%%%%%%%%

\subsection{Convolutional information processing on graphons}

The connection between graphs and graphons prompts a natural link between convolutional signal processing on graphs and convolutional signal processing on graphons. We begin by recalling that in graphon signal processing (Gphon-SP) a signal on a graphon $W(x,y)$ is given by a pair $(W,\boldsymbol{x})$ where $\boldsymbol{x}$ is an element of $L^{2}([0,1])$.

In the same way that graphons can be induced by graphs, some graphon signals can be induced by graph signals. To see this, let us consider the graph signal $(G,\bbx)$ and the graphon signal $(W_G,\boldsymbol{x})$ where $W_G$ is induced by $G$. Then, we can say that $(W_G,\boldsymbol{x})$ is induced by $(G,\bbx)$ if
%
%
%
%\begin{equation}\label{eq_wsig_from_gsig}
$
\boldsymbol{x}(t)
                  =
                    \mathsf{step}\left( \bbx \right)
                  =    
                    \bbx ( \lceil tN \rceil )
,
$
%\end{equation}
%
%
%
where $t\in [0,1]$. This is, graphon signals induced by graph signals are piece-wise constant functions defined on a regular partition of $[0,1]$. Notice that a graphon signal is a particular case of a \textit{node level statistic} as defined in~\cite{Diao2016ModelfreeCO}.

To define convolutions for graphon signals, it is necessary to first introduce a shift operator~\cite{Pschel2006AlgebraicSP,alejo_algnn_j,alejo_algnn_c}. Following~\cite{lovaz2012large}, we define the graphon shift operator $\boldsymbol{T}_W$ on a graphon $W(x,y)$ acting on a graphon signal $(W,\boldsymbol{x})$ by
\begin{equation}
               \left(
                      \boldsymbol{T}_{W}\boldsymbol{x}
               \right)(v)
                            =
                               \int_{0}^{1}W(u,v)\boldsymbol{x}(u)du
                               .
\end{equation}
Notice that $\boldsymbol{T}_W$ is Hilbert-Schmidt~\cite{lovaz2012large}. Now, we can define convolutional filters for graphon signals using polynomial operators written in terms of $\boldsymbol{T}_W$. The convolutional filtering of a graphon signal $(W,\boldsymbol{x})$ by means of a graphon filter $h(\boldsymbol{T}_W)=\sum_{k=0}^{K}h_{k}\boldsymbol{T}_{W}^{k}$ is given by
\begin{equation}
h(\boldsymbol{T}_{W})\boldsymbol{x}
                           =
                              \sum_{k=0}^{K}h_{k}\boldsymbol{T}_{W}^{k} \boldsymbol{x}, 
\end{equation}
where $\boldsymbol{T}_{W}^{k}$ represents the $k$-times composition of  $\boldsymbol{T}_{W}$. It is important to remark that the graphon filter $h(\boldsymbol{T}_W)$ can be characterized by means of the scalar polynomial~\cite{alejo_algnn_j, alejo_algnn_c},
\begin{equation}
   h(t) = \sum_{k=0}^{K}h_{k}t^{k}.
\end{equation}
In what follows we will refer to $h(t)$ as the polynomial or functional representation of the graphon filter $h(\boldsymbol{T}_W)=\sum_{k=0}^{K}h_{k}\boldsymbol{T}_{W}^{k}$. The representation $h(t)$ is also known as the frequency representation of the graphon filter~\cite{alejo_algnn_j, alejo_algnn_c}.

%%%%%%%%%%%%%%%%%%%%%%%%%%%%%%%%%%%%%%%
%%%%%%%%%%%% Sub-Sub-Section %%%%%%%%%%%%%%
%%%%%%%%%%%%%%%%%%%%%%%%%%%%%%%%%%%%%%%

%\subsubsection{Induced graphon filtering}

The relationship between graph signals and their induced graphon counterparts translate to the action of the shift operators and the filters. However, as we will indicate in the theorem below, that transference requires a scaling.

%%-------------------------------------------
%%---------------- THEOREM ------------------
%%-------------------------------------------

\begin{theorem}\label{thm_induced_gphon_filtering}

Let $(W_{G} , \boldsymbol{x})$ be a graphon signal induced by the graph signal $(G,\bbx)$. Let $h(t)=\sum_{k=0}^{K-1}h_{k}t^k$ be a filter and $\boldsymbol{y} = h\left( \boldsymbol{T}_{W_G} \right)\boldsymbol{x}$, where $\boldsymbol{T}_{W_G}$ is the graphon shift operator in $W_G$. Then, it follows that
\begin{equation}
\boldsymbol{y}
          =
          \mathsf{step}
          \left(
           h\left( 
                 \frac{\bbS_G}{\vert V(G)\vert}
           \right)\bbx
          \right) 
,
\end{equation}
where $\bbS_G$ is the shift operator on $G$.
\end{theorem}

\begin{proof} See Appendix~\ref{proof_thm_induced_gphon_filtering} \end{proof}

%%----------- End of Theorem ----------------

From Theorem~\ref{thm_induced_gphon_filtering} we observe that in order to transfer the filtering operation between graph signals and their induced graphon counterparts, we have to scale the shift operator in the graph according to $\bbS_G \to \bbS_{G}/\vert V(G)\vert$. Notice that although $\boldsymbol{y}\neq  \mathsf{step}( h(\bbS_G)\bbx )$, it is possible to determine uniquely $\boldsymbol{y}$ from $\mathsf{step}( h(\bbS_G)\bbx )$ given the knowledge of $h(t)$. This implies that it is possible to learn the coefficients of a filter on the graph and transfer them to the induced graphon. Likewise, it is possible to characterize the properties of the filters on the induced graphon operators and transfer such properties to the filters on the graph. We will exploit this relationship in order to analyze GNNs. More specifically, we characterize the properties of the graph and GNN operators on their induced graphon representations.

%%%%%%%%%%%%%%%%%%%%%%%%%%%%%%%%%%%%%%%%%%%
%%%%%%%%%%%%%% SUB-SECTION %%%%%%%%%%%%%%%%
%%%%%%%%%%%%%%%%%%%%%%%%%%%%%%%%%%%%%%%%%%%

\subsection{Graphon Neural Networks (Gphon-NNs)}

Convolutional architectures can be established in multiple domains including graphons~\cite{alejo_algnn_j}. This is, information processing on graphons can be combined with pointwise nonlinearity operators to obtain \textit{graphon neural networks (Gphon-NNs)}~\cite{alej2020graphon}. Formally, a Gphon-NN is a stacked layered structure analogous to a GNN -- see Fig.~\ref{fig:GNNbasicfig} -- where the convolutions are carried out by graphon filters on graphon signals. In the $i$-th layer of a Gphon-NN an input signal $\boldsymbol{x}_{i-1}$ is filtered to obtain $\boldsymbol{y}_i = \sum_{k=0}^{K}h_{i,k}\boldsymbol{T}_{W_i}^{k}\boldsymbol{x}_{i-1}$. Then, $\boldsymbol{y}_i$ is transformed by a pointwise nonlinearity $\eta_i$ -- assumed to be Lipschitz -- to obtain $\boldsymbol{z}_i = \eta_i (\boldsymbol{y}_i)$ --, and right after a pooling operator generates $\boldsymbol{x}_i = P_i \left(\boldsymbol{z}_i \right)$. The operator $P_i$ reduces the complexity and or degrees of freedom of the information in $\boldsymbol{z}_i$ while preserving essential features. Additionally, $P_i$ can be embedded in the graphon shift operator associated to each layer. Then, the output of the $i$-th layer of a Gphon-NN can be written as
%
%
%
%\begin{equation}\label{eq:HinGphon-nn}
$
\boldsymbol{x}_{i}
           =
                    \eta_i \left( 
                                     \boldsymbol{H}_i (\boldsymbol{T}_{W_i}) \boldsymbol{x}_{i-1}
                              \right)
                    ,
                    \;
                    \text{with}
                    \;
                    \boldsymbol{H}_i (\boldsymbol{T}_{W_i}) = \sum_{k=0}^{K}h_{i,k}\boldsymbol{T}_{W_i}^{k}
                    .
$
%\end{equation}
%
%
%
This expression can be extended trivially to multiple features but such extension is not central to our analysis. 

Similar to the scenario of GNNs, the coefficients $h_{i,k}$ are learned from the input data. Given a training set $\mathcal{T}=\{(\boldsymbol{x},\boldsymbol{y})\}$ with inputs $\boldsymbol{x}$ and outputs $\boldsymbol{y}$, the Gphon-NN learns a representation that relates an output $\hat{\boldsymbol{y}}$ to a given input $\hat{\boldsymbol{x}}$ with $(\hat{\boldsymbol{x}},\hat{\boldsymbol{y}})\notin\mathcal{T}$. We will represent the mapping operator of a Gphon-NN with $L$ layers by $\mathbf{\Phi}\left( \boldsymbol{x}, \{ \ccalF_\ell \}_{\ell=1}^{L}, \{ \boldsymbol{T}_{W_\ell} \}_{\ell=1}^{L}\right)$. The input signal to the Gphon-NN is $\boldsymbol{x}$, while $ \ccalF_\ell$ and $\boldsymbol{T}_{W_\ell}$ are the sets that indicate the properties of the filters and the graphon shift operators in the layer $\ell$, respectively.

%%-------------------------------------
%%------------ REMARK -----------------
%%-------------------------------------

\begin{remark}\normalfont
\label{rmk_gphon_from_gnn}

Notice that as a consequence of the unique relationship between graphs and their induced graphons, it is possible to use Gphon-NNs to process information on GNNs. To see this, let us consider the GNN with graph layers given by $\{ G_\ell \}_{\ell=1}^{L}$. Then, there is an induced Gphon-NN with graphon layers given by $\{ W_{G_\ell} \}_{\ell=1}^{L}$. Let $(W_{G_\ell},\boldsymbol{x})$ be the graphon signal induced from $(G_{\ell},\bbx)$. Then, the processing of $(G_\ell, \bbx)$ in the $\ell$-th layer of the GNN can be carried out by processing the graphon signal $(W_{G_\ell},\boldsymbol{x})$ in the $\ell$-th layer of the induced Gphon-NN -- taking into account the scaling given in Theorem~\ref{thm_induced_gphon_filtering}. Then, there is a one to one relationship between $\mathbf{\Phi}\left( \bbx, \{ \ccalF_\ell \}_{\ell=1}^{L}, \{ \bbS_\ell \}_{\ell=1}^{L}\right)$ and $\mathbf{\Phi}\left( \boldsymbol{x}, \{ \ccalF_\ell \}_{\ell=1}^{L}, \{ \boldsymbol{T}_{W_{G_\ell}} \}_{\ell=1}^{L}\right)$. Additionally, if the graphs $\{ G_\ell \}_{\ell=1}^{L}$ are obtained from $W(x,y)$ by a pooling method, the effects of pooling on the GNN can be studied analyzing  the term
$
\left\Vert 
\mathbf{\Phi}\left( \boldsymbol{x}, \{ \ccalF_\ell \}_{\ell=1}^{L},  \boldsymbol{T}_{W} \right)
-
\mathbf{\Phi}\left( \boldsymbol{x}, \{ \ccalF_\ell \}_{\ell=1}^{L}, \{ \boldsymbol{T}_{W_{G_\ell}} \}_{\ell=1}^{L}\right)
\right\Vert_{2}.
$
This means that the effects of the pooling operation can be analyzed by means of a comparison between the Gphon-NN induced from the GNN, and a Gphon-NN defined on the graphon $W(x,y)$ -- where $W(x,y)$ is associated to the graph in the first layer of the GNN. In the following section we leverage these facts to study the effects of the graphon pooling methods proposed. We end this remark emphasizing that the unique relationship and equivalence between the operations of filtering in graphs and their induced graphons, is indeed an analytical equivalence that has to be taken with caution when performing numerical operations in both domains. This is a consequence of the fact that the scaling -- see Theorem~\ref{thm_induced_gphon_filtering} -- given by $\bbS_{G} \to \bbS_{G} / \vert V(G)\vert$ may affect severely the condition of the matrix operators when $\bbS_G$ is sparse. 

\end{remark}

%%--------- End of Remark -------------

%% file: v16/sec_graphonPool.tex
%!TEX root =../compiler_paper.tex

%%%%%%%%%%%%%%%%%%%%%%%%%%%%%%%%%%%%%%%%%%%%%
%%%%%%%%%%%%%%%% SECTION %%%%%%%%%%%%%%%%%%%%
%%%%%%%%%%%%%%%%%%%%%%%%%%%%%%%%%%%%%%%%%%%%%

\section{Graphon Pooling}\label{sec_graphon_pooling}

%%-----------------------------------
%%---------- FIGURE ------------------
%%-----------------------------------

\begin{figure}
	\centering\includegraphics[width=0.5\textwidth]{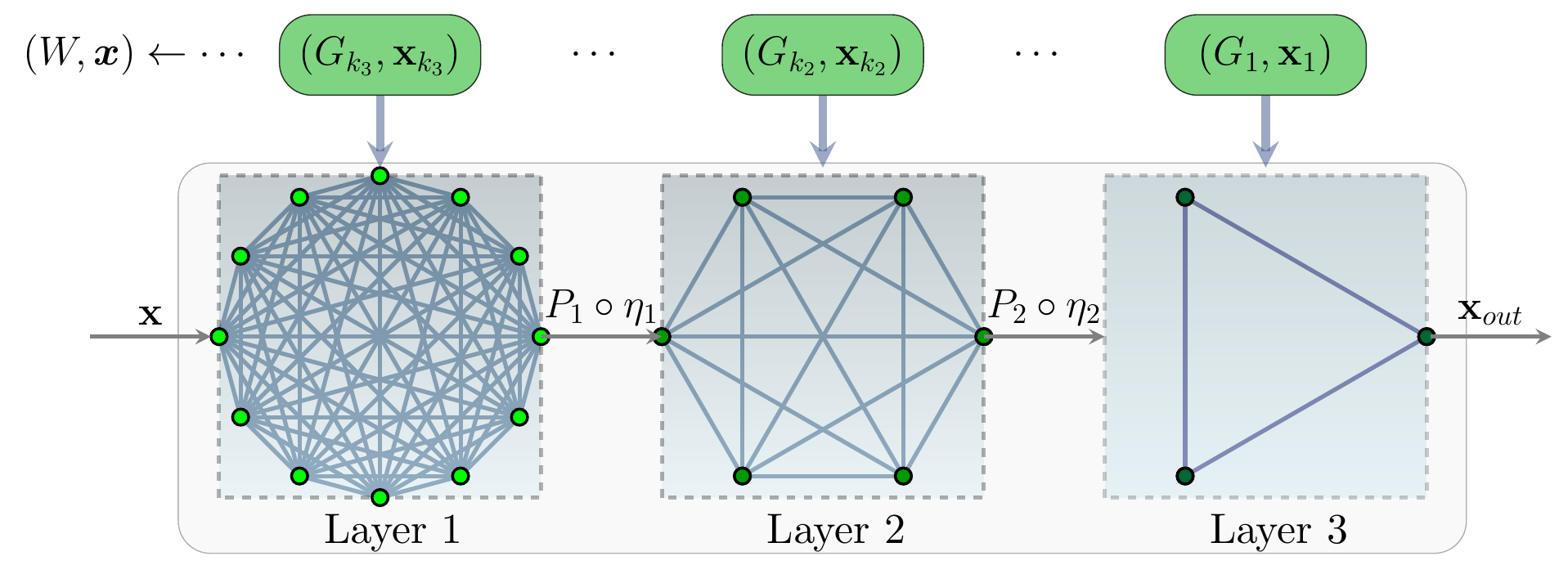}
 \caption{Schematic representation of the graphon pooling concept. The sequence of graph signals $\displaystyle \left( G_i , \mathbf{x}_i \right)$ converge to the graphon signal $\displaystyle \left( W, \boldsymbol{x} \right)$. Then, a finite subsequence of graph signals $\displaystyle\left\lbrace (G_1 ,\bbx_1), (G_{k_2},\bbx_{k_2}), (G_{k_3},\bbx_{k_3})\right\rbrace$ is selected to define the layers of the GNN, where the largest graph is associated to the first layer. This process specifies in an implicit way the action of the pooling operators $P_i$ between layers since the convergence of $(G_i,\bbx_i)$ guarantees some minimum common structural properties for all the elements in the sequence.}
 \label{fig:gnnfraomgseq}
 \end{figure}

%%-------- End of Figure ----------------

The fundamental idea of the graphon pooling approach proposed in this paper relies on leveraging subsequences of a convergent sequence of graph signals to build a GNN -- see Fig.~\ref{fig:gnnfraomgseq} . The largest graph of such subsequence is associated to the first layer of the GNN, while the smaller graphs are associated to subsequent layers. The convergence of the graph signals implies spectral consistency on the graphs -- that are also convergent -- and that there is functional convergence of the information associated to the nodes of the graphs. The sequences are built from the data embedded in a graphon representation.

It is important to remark that in practice one starts with the data defined on a large graph and not a graphon. However, as we have shown that every graph has a natural induced graphon representation, we will show in subsection~\ref{sub_sec_consistency_gphonpooling} that such induced graphon representation is a good approximation of the graphon limit if the graph is large enough. This indicates that the original graph provides naturally the graphon used to build the elements of the convergent graph sequence. In what follows we describe the numerical approaches used to build such sequences.

%%-------------------------------
%%--------- Method 1 --------------
%%-------------------------------

\begin{figure}
	\centering\includegraphics[angle=0,scale=0.5]{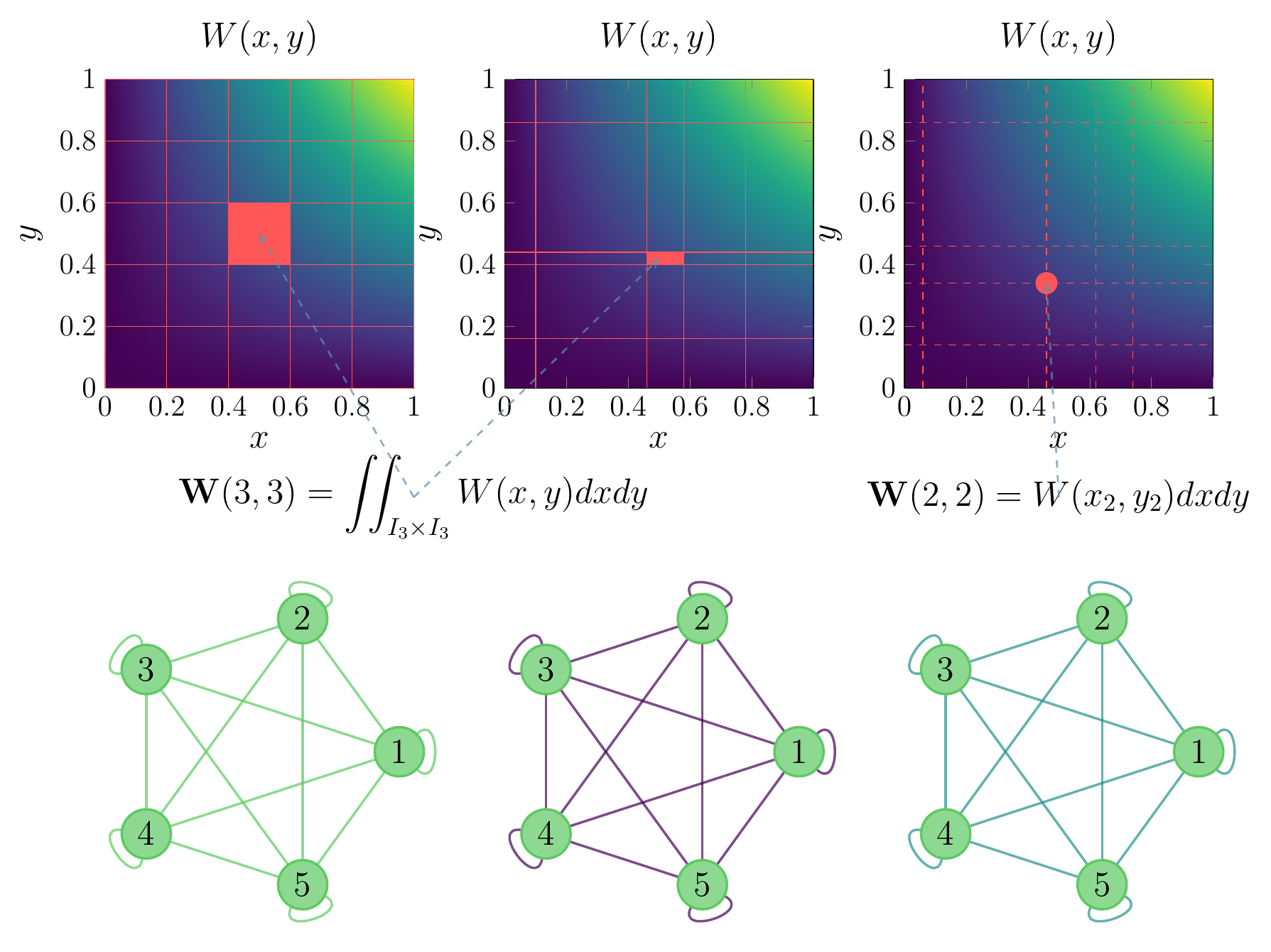}
	\caption{Generation of a graph adjacency matrix $\bbA^{G}$ from an underlying Graphon $W(x,y)$ by means of methods \textbf{M1} (left), \textbf{M2} (center), and \textbf{M3} (right). The grids in the first row (left and center) define partitions of $[0,1]^2$, while the dotted lines in the right (first row) indicate values of $x$ and $y$ where $W(x,y)$ is evaluated. Left: discretization method \textbf{M1} where a \textit{regular} grid is used to build $\bbA^{G}$, and each entry of $\bbA^{G}$ is given by the volume below $W(x,y)$ in each element of the partition. Center: discretization method \textbf{M2}, where the subdomains of an  \textit{irregular} grid are used to obtain the entries of $\bbA^{G}$. Right: discretization method \textbf{M3} where each entry of $\bbA^G$ is obtained evaluating $W(x,y)$ at an specific point $(x,y)\in [0,1]^2$.} 
	\label{fig:disctmethods}
\end{figure}

\begin{method}[\textbf{M1}, regular integration]\label{disctmethod:regint}\normalfont

In this approach we start by considering a graphon $W(u,v)$. We then build a sequence of graphs $\{G_{\ell}\}_{\ell}\rightarrow W(u,v)$ using a uniform partition of $[0,1]^2 \subset \mbR^2$ -- see Fig.~\ref{fig:disctmethods} (left). We associate the entries of the adjacency matrix of the graph with the volume below $W(u,v)$ in each of the elements of the partition. Then, the adjacency matrix $\mathbf{A}^{G_{\ell}}$ for each graph $G_{\ell}$ is given by

\begin{equation} \label{eqn:discregintegration}
\mathbf{A}^{G_{\ell}} (i,j)
                         =
   \frac{1}{\Delta_{i,j}}\int_{\rho(j)}^{\rho(j+1)}\int_{\rho(i)}^{\rho(i+1)}W(u,v)dudv
                        ,
\end{equation}
with $\rho (i)=(i-1)/N_{\ell}$ for all $i\in\{ 1,2,\ldots N_{\ell} \}$, $\Delta_{i,j} = 1/N_\ell^2$, where $N_{\ell}$ is the number of nodes in $G_{\ell}$, and $\rho(N_\ell +1)=1$. The schematic representation of this method is depicted in Fig.~\ref{fig:disctmethods}~(left).

\end{method}

%%-------------------------------
%%--------- Method 2 --------------
%%-------------------------------

\begin{method}[\textbf{M2}, irregular integration]\label{disctmethod:irrint}\normalfont

In this approach we build the graphs in the sequence considering an irregular partition of $[0,1]^2 \subset \mbR^2$ -- see Fig.~\ref{fig:disctmethods} (center). We associate each entry of the adjacency matrix to the volume below $W(x,y)$ in each element of the partition. To build the partition we use a random uniform distribution over $[0,1]$. The adjacency matrix $\mathbf{A}^{G_{\ell}}$ of each $G_{\ell}$ is given by \eqref{eqn:discregintegration}, where $\Delta_{i,j}$ is the area of the partition element $[\rho(j),\rho(j+1)]\times [\rho(i),\rho(i+1)]$, and the map $\rho$ is defined according to a random uniform distribution in $[0,1]$ with the following restrictions: $\rho(1)=0$, $\rho(i)\leq\rho(i+1)$, and $\rho(N_\ell +1)=1$. This procedure is illustrated in Fig.~\ref{fig:disctmethods} (center).

\end{method}

%%-------------------------------
%%--------- Method 3 --------------
%%-------------------------------

\begin{method}[\textbf{M3}, irregular sampling]\label{disctmethod:irrsamp}\normalfont
	
In this approach we obtain the adjacency matrix of the graphs in the sequence sampling  values of an underlying graphon $W(x,y)$. To perform the sampling we use a random uniform distribution to select a set of $N_{\ell}$ points in $[0,1]$. The relationship between the nodes and the points in $[0,1]$ is defined by a map $\rho$ with $\rho(i)\leq\rho(i+1)$. Then, the adjacency matrix $\mathbf{A}^{G_{\ell}}$ of the graph $G_{\ell}$ is given by
\begin{equation}\label{eqn:disctrandsamp}
\mathbf{A}^{G_{\ell}}(i,j)
                    =
                         W\left(\rho(i),\rho(j)\right).
\end{equation}
The discretization method \textbf{M3} is illustrated in Fig.~\ref{fig:disctmethods} (right).
\end{method}

%%-------------------------------------
%%------------ FIGURE -------------------
%%-------------------------------------

%%---------- End of Figure ----------------

%%%%%%%%%%%%%%%%%%%%%%%%%%%%%%%%%%%%%%%%%%
%%%%%%%%%%%%% Sub-Section %%%%%%%%%%%%%%%%
%%%%%%%%%%%%%%%%%%%%%%%%%%%%%%%%%%%%%%%%%%

\subsection{Labeling and mapping of signal components}

The labeling assigned to the graph $\bbA^G$ generated by means of \textbf{M1}, \textbf{M2}, and \textbf{M3} is inherited from the natural ordering associated to the interval $[0,1]$. To see this we recall that for all the discretization methods proposed we have $\rho(i)<\rho(i+1)$. Then, the index $i$ naturally \textit{labels} the nodes in $\bbA^G$ and the components of any graph signal defined on $\bbA^G$.

Methods \textbf{M1}, \textbf{M2}, and \textbf{M3} do not establish a correspondence between components of signals on graphs of different sizes. Since \textbf{M1}/\textbf{M2}/\textbf{M3} are used to define the layers of a GNN, a consistent mapping between the signal components is necessary. We perform such mapping leveraging the associated induced graphon signals and interpolation. The details of this procedure are presented in the following paragraphs.

%%-------------------------------------
%%------ Signal Mapping with M1 -------
%%-------------------------------------

\vspace{3mm}
\noindent\textbf{Signal mapping with M1/M2}: In order to perform the interpolation of signals when \textbf{M1}/\textbf{M2} is used, we take as a reference the center point of the intervals defining the partition used to build the integration grid. Let us consider the pooling operation on the graph $G_{\ell}$ to obtain $G_{\ell+1}$. We denote by $I_{i}^{(\ell)}$ the intervals of the partition of $[0,1]$ associated to $W_{G_\ell}$ and by $\overline{I_{i}^{(\ell)}}$ the center point of $I_{i}^{(\ell)}$. Now, we take into account that for any $I_{j}^{(\ell+1)}$ there exists $I_{i}^{(\ell)}$ and $I_{i+1}^{(\ell)}$ such that

\begin{equation}
\overline{I_{i}^{(\ell)}} 
<  
\overline{I_{j}^{(\ell+1)}} 
<
\overline{I_{i+1}^{(\ell)}}
.
\end{equation}
Then, if the graphs $G_\ell$ are obtained from $W(x,y)$ using \textbf{M1}/\textbf{M2} the signal $(W_{G_\ell},\boldsymbol{x}_\ell)$ is mapped to the signal $(W_{G_{\ell+1}},\boldsymbol{x}_{\ell+1})$, where

\begin{equation}
\boldsymbol{x}_{\ell+1}(u)
    =
     \frac{1}{2}
     \left(
          \boldsymbol{x}_{\ell}\left( I_{i}^{(\ell)}\right)
          +
          \boldsymbol{x}_{\ell}\left( I_{i+1}^{(\ell)}\right)
     \right)
      ,
     \quad 
        u\in I_{j}^{(\ell+1)} 
        .
\end{equation}
Since $\boldsymbol{x}_{\ell}$ is piece-wise constant on the intervals $I_{i}^{(\ell)}$, the value $\boldsymbol{x}_{\ell}\left( I_{i}^{(\ell)}\right)$ is well defined.

%%------- End of mapping in M1/M2 --------

%%-------------------------------------
%%------ Signal Mapping with M3 -------
%%-------------------------------------

\vspace{3mm}
\noindent\textbf{Signal mapping with M3}: For the interpolation using \textbf{M3} we use the location of the sampling points. We perform the pooling operation on the graph $G_{\ell}$ to obtain $G_{\ell+1}$ and we denote by $t_{i}^{(\ell)}$ the sampling points in $[0,1]$ associated to $W_{G_\ell}$. We recall that for any $t_{j}^{(\ell+1)}$ there exists $t_{i}^{(\ell)}$ and $t_{i+1}^{(\ell)}$ such that

\begin{equation}
t_{i}^{(\ell)} 
<  
t_{j}^{(\ell+1)} 
<
t_{i+1}^{(\ell)}
.   
\end{equation}
Therefore, if the graphs $\{ G_\ell \}_\ell$ are obtained from $W(x,y)$ by means of method \textbf{M3} the signal $(W_{G_\ell},\boldsymbol{x}_\ell)$ is mapped to the signal $(W_{G_{\ell+1}},\boldsymbol{x}_{\ell+1})$, where

\begin{equation}
\boldsymbol{x}_{\ell+1}\left( t_{j}^{(\ell+1)} \right)
    =
     \frac{1}{2}
     \left(
          \boldsymbol{x}_{\ell}\left( t_{i}^{(\ell)}\right)
          +
          \boldsymbol{x}_{\ell}\left( t_{i+1}^{(\ell)}\right)
     \right)
        .
\end{equation}
%
%
%

%%------- End of mapping in M3 --------

%%-------------------------------------
%%------------ REMARK -----------------
%%-------------------------------------

\begin{remark}\normalfont

Notice that among the pooling methods presented, \textbf{M3} exhibits the lowest computational complexity, and it is indeed the simplest approach to perform pooling. However, as we will show in the following sections, although \textbf{M1} and \textbf{M2} have higher computational cost, they are endowed with structural properties that allow the derivation of concrete error performance bounds and allow approximate label permutation invariance/equivariance. 

\end{remark}

%%---------- End of Remark ------------

%%%%%%%%%%%%%%%%%%%%%%%%%%%%%%%%%%%%%%%%%%%%
%%%%%%%%%%%%%% Sub - Section %%%%%%%%%%%%%%% 
%%%%%%%%%%%%%%%%%%%%%%%%%%%%%%%%%%%%%%%%%%%%

\subsection{Consistency of graphon pooling}
\label{sub_sec_consistency_gphonpooling}

In this section we elaborate about the theoretical foundations of graphon pooling. We start with introducing a result providing the convergence guarantees for those sequences of graphs obtained from a graphon using the discretization methods introduced before. In what follows we will refer to discretization methods \textbf{M1}, \textbf{M2} and \textbf{M3} as the pooling methods.

%%-----------------------------------------
%%---------------- THEOREM ----------------
%%-----------------------------------------

\begin{theorem}\label{thm_M1andM2_convergence}
	Let $\{ G_\ell \}_\ell$ be a sequence of graphs generated from the graphon $W(x,y)$ by the discretization methods \textbf{M1} and \textbf{M2}. If $\{ W_{G_\ell} \}_\ell$ is the sequence of induced graphons associated to $\{ G_\ell \}_\ell$, then
	$\{ W_{G_\ell} \}_\ell\rightarrow W$ almost everywhere with	
	\begin{equation}\label{eq_thm_M1andM2_convergence}
	  \left\Vert 
	        \boldsymbol{T}_{W}
	        -
	        \boldsymbol{T}_{W_{G_\ell}}
	  \right\Vert_{\infty\rightarrow 1}  
	                   \leq 
	                    \frac{8}{\sqrt{\log \vert V(G_\ell ) \vert}}
                     .
	\end{equation}
\end{theorem}
\begin{proof}
	See Appendix~\ref{proof_thm_M1andM2_convergence}.
\end{proof}

%%----------- End of Lemma ----------------

Theorem~\ref{thm_M1andM2_convergence} provides the guarantees of convergence of a sequence of graphs used to build a large GNN. This result at the same time assures the spectral structural consistency of the graphs used in the GNN. We emphasize that Theorem~\ref{thm_M1andM2_convergence} is an immediate consequence of the fact that kernels can be approximated well by step functions in the $L_1$ norm -- see Appendix~\ref{proof_thm_M1andM2_convergence}. 

%This is also formalized in~\cite{lovaz2012large}~(Proposition 9.8).

In graphon pooling the assumption that there is a closed form expression for the limit graphon can indeed be relaxed when considering a piecewise constant representation of the largest finite graph in the sequence. This is not just an approximation, but as we will show in the following lemma it is indeed a natural way to represent a graphon with zero error in terms of the cut metric.

%%-------------------------------------------
%%---------------- THEOREM ------------------
%%-------------------------------------------

\begin{lemma}[Adapted from Lemma 9.11 in~\cite{lovaz2012large}]\label{lemma_cutmetric_zeroerror_step}
	
Let $G$ be a graph obtained from the graphon $W(x,y)$ using the discretization methods $\textbf{M1}$ and $\textbf{M2}$. Then, there exists a graph $H$ with $\vert V(H)\vert\leq 4 \vert V(G) \vert$ such that

\begin{equation}
\left\Vert 
W
-
W_{G}
\right\Vert_{\square}
=
\left\Vert 
W_{H}
-
W_{G}
\right\Vert_{\square}   
,           
\end{equation}
where $H$ is obtained from $W(x,y)$ using methods $\textbf{M1}$ and $\textbf{M2}$ with a refined partition of $W_G$. The terms $W_G$ and $W_H$ are the graphons induced by $G$ and $H$, respectively. 
	
\end{lemma}

%%----------- End of Theorem ----------------

Lemma~\ref{lemma_cutmetric_zeroerror_step} highlights fundamental properties of the graphon pooling methods proposed in this paper. In particular, it remarks that in terms of the cut norm a step function representation of any graphon is enough to capture structural properties. In practice this points to the fact that a large graph induces a graphon that is a good approximation of the graphon limit. Then, if a finite sequence of graphs in a GNN is obtained from a graphon, one could consider that the largest graph in the sequence can induce a graphon that is indistinguishable from the graphon limit -- in terms of the cut norm. It is equally important to highlight that the partition associated to $W_{H}$ has at most $4\vert V(G)\vert$ elements. In practice this implies that one can reduce the size of a step function graphon up to a factor of 4 and still preserve the same distance with respect to the graphon limit.

Taking into account the cut norm distance between the induced graphons of a convergent sequence $\{ W_{G_\ell} \}_{\ell}$ and the graphon limit $W(x,y)$, we can obtain upper bounds for the change of the filter outputs when implemented on $W_{G_\ell}$ and $W$. We state this formally in the following theorem.

%%-------------------------------------------
%%---------------- THEOREM ------------------
%%-------------------------------------------

\begin{theorem}\label{thm_change_filters}

Let $h(t) = \sum_{k=0}^{\infty}h_{k}t^{k}$ be the functional representation of a graph/graphon filter with Lipschitz constant $C$. Let $G$ be a graph obtained from the graphon $W(x,y)$ using methods \textbf{M1}, \textbf{M2} or \textbf{M3}. Let $\Vert \boldsymbol{T}_W - \boldsymbol{T}_{W_G} \Vert_{HS}\leq \gamma \Vert \boldsymbol{T}_W - \boldsymbol{T}_{W_{G}} \Vert_2 $ for $\gamma>0$, where $\boldsymbol{T}_{W}$ and $\boldsymbol{T}_{W_{G}}$ are the graphon shift operators of $W$ and $W_{G}$ respectively. Then, it follows that
\begin{equation}\label{eq_thm_change_filters_1}
\left\Vert 
               h\left(\boldsymbol{T}_W \right)
                 -
               h\left(\boldsymbol{T}_{W_G}\right)
\right\Vert_2                 
               \leq
               \Omega \left( \bbT_{W} \right)
                +
               \ccalO\left( \Vert
                                  \boldsymbol{T}_{W}-\boldsymbol{T}_{W_{G}}
                               \Vert_{2}^2 
                        \right)
                ,
\end{equation} 
where 
\begin{equation}\label{eq_thm_change_filters_2}
\Omega\left( 
           \bbT_{W} 
       \right)  
       =
                        \gamma
                 C
\sqrt{8                 
\left\Vert
            W - W_{G}
\right\Vert_{\square} 
}
                 .
\end{equation}
Additionally, if $G$ is obtained by \textbf{M1}/\textbf{M2}, there exists a graph $H$ obtained from $W(x,y)$ using \textbf{M1}/\textbf{M2} with a refinement of the partition used in $W_{G}$ and with $\vert V(H) \vert \leq 4\vert V(G)\vert$ such that	
\begin{equation}\label{eq_thm_change_filters_3}
\Omega\left( 
           \bbT_{W} 
       \right)  
       =    
        \gamma
         C
\sqrt{8                 
\left\Vert
            W_{H} - W_{G}
\right\Vert_{\square} 
}
                 .       
\end{equation}
$\Vert \cdot \Vert_2$ represents the norm $\Vert \cdot \Vert_{2\rightarrow 2}$ and $\Vert \cdot \Vert_{HS}$ indicates the Hilbert-Schmidth norm.
\end{theorem}

\begin{proof}
See Appendix~\ref{sec_proof_thm_change_filters}.
\end{proof}

%%------------- End of Theorem ----------------

Theorem~\ref{thm_change_filters} and~(\ref{eq_thm_change_filters_2}) show an upper bound, $\Omega (\boldsymbol{T}_{W})$, of the changes of a filter $h(t)$ when instantiated in the graphon $W(x,y)$ and an induced graphon $W_{G}$ from the graph $G$ obtained from $W(x,y)$ by means of \textbf{M1}/\textbf{M2}/\textbf{M3}. Somewhat expected is the dependency of $\Omega (\boldsymbol{T}_{W})$ of the smoothness of $h(t)$, measured by its Lipschitz constant $C$. On the other hand, a more interesting feature of $\Omega (\boldsymbol{T}_{W})$ is its dependency with respect to the square root of the distance between $W(x,y)$ and $W_{G}$ in terms of the cut norm.

Note that~(\ref{eq_thm_change_filters_2}) and~(\ref{eq_thm_change_filters_3}) highlight substantial differences between \textbf{M1}, \textbf{M2} and \textbf{M3}. First, we can see that by means of \textbf{M1}/\textbf{M2} it is possible to reduce up to a factor of four the number of nodes when performing pooling while at the same time keeping structural properties associated to the underlying graphon. This is reflected by the fact that the value of $\Omega (\boldsymbol{T}_{W})$ is explicitly calculated using the graph $H$ -- obtained with \textbf{M1}/\textbf{M2} -- with $\vert V(H)\vert \leq 4 \vert V(G)\vert$. This guarantee does not take place for \textbf{M3}. Second, it is important to remark that for~(\ref{eq_thm_change_filters_3}) to be valid we must guarantee that the application of \textbf{M1}/\textbf{M2} takes place considering \textit{refined partitions}. This is, the partition used to generate $H$ from $W(x,y)$ is a refined partition of the partition used to generate $G$. This in itself points to an advantage of \textbf{M1} over \textbf{M2}. More specifically, notice that the partitions used for \textbf{M1} will be automatically refinements as long as the pooling operation reduces the number of nodes by an integer factor, while the random geometric nature of \textbf{M2} will most likely not lead to partitions that are refined versions of each other. 

With Theorem~\ref{thm_change_filters} at hand, we now establish an upper bound for the change of the GNN and Gphon-NN operators when considering graphs obtained from a given graphon using methods \textbf{M1}, \textbf{M2} and \textbf{M3}.

%%-------------------------------------------------------------
%%-------------- THEOREM STABILITY Graphon NN -----------------
%%-------------------------------------------------------------

\begin{theorem}\label{theorem:stabilityAlgNN0}

Let $\mathbf{\Phi}\left(\boldsymbol{x}, \{ \mathcal{F}_{\ell}\}_{\ell=1}^{L},\boldsymbol{T}_{W}\right)$ be the mapping operator of a Gphon-NN where no pooling operation is performed. Let $\mathbf{\Phi}\left(\boldsymbol{x}, \{ \mathcal{F}_{\ell} \}_{\ell=1}^{L},\{ \boldsymbol{T}_{W_{G_\ell}} \}_{\ell=1}^{L} \right)$ be the mapping operator of a Gphon-NN where each layer is defined by the graphon $W_{G_\ell}$, where the $G_\ell$ are generated from $W(x,y)$ by methods \textbf{M1}/\textbf{M2}/\textbf{M3}. Let $\Vert \boldsymbol{T}_W - \boldsymbol{T}_{W_{G_\ell}} \Vert_{HS}\leq \gamma \Vert \boldsymbol{T}_W - \boldsymbol{T}_{W_{G_\ell}} \Vert_2 $ for $\gamma>0$, where $\boldsymbol{T}_{W}$ and $\boldsymbol{T}_{W_{G_\ell}}$ are the graphon shift operators of $W$ and $W_{G_\ell}$, respectively. Then, it follows that
\begin{multline}\label{eq_thm_stabilityAlgNN0_0}
	\left\Vert
	\mathbf{\Phi}\left(\boldsymbol{x},\{ \mathcal{F}_{\ell} \}_{\ell=1}^{L},\boldsymbol{T}_{W}\right)-
	\mathbf{\Phi}\left(\boldsymbol{x}, \{ \mathcal{F}_{\ell} \}_{\ell=1}^{L},\{ \boldsymbol{T}_{W_{G_\ell}} \}_{\ell=1}^{L} \right)
	\right\Vert_2
	\\
	\leq
\Omega\left( \boldsymbol{T}_{W} \right)
            \left\Vert
                   \boldsymbol{x}
            \right\Vert_{2}
                 +
                 \ccalO
                    \left( \Vert
                             \boldsymbol{T}_{W}-\boldsymbol{T}_{W_{G_\ell}}
                           \Vert_{2}^2 
                    \right)
	,
\end{multline}
where
\begin{equation}
\Omega\left( 
         \boldsymbol{T}_{W} 
       \right)
       =
       \sqrt{8}C\gamma
	\left(
	\sum_{\ell=1}^{L}
\sqrt{                 
\left\Vert
            W - W_{G_\ell}
\right\Vert_{\square} 
}
      \right)
      .
\end{equation}
Additionally, if the $G_{\ell}$ are generated by \textbf{M1}/\textbf{M2}, there exist graphs $H_{\ell}$ obtained from $W(x,y)$ using \textbf{M1}/\textbf{M2} with a refinement of the partitions used in $W_{G_{\ell}}$ and with $\vert V(H_{\ell}) \vert \leq 4\vert V(G_{\ell})\vert$ such that	
\begin{equation}
\Omega\left( 
         \boldsymbol{T}_{W} 
       \right)
       =
       \sqrt{8}C\gamma
	\left(
	\sum_{\ell=1}^{L}
\sqrt{                 
\left\Vert
            W_{H_\ell} - W_{G_\ell}
\right\Vert_{\square} 
}
      \right)
      .
\end{equation}
Here $\mathcal{F}_{\ell}$ is a set of $C$-Lipschitz filters and the index $(\ell)$ makes reference to quantities and constants associated to the $\ell$-th layer.

\end{theorem}

\begin{proof}
See Appendix~\ref{prooftheorem:stabilityAlgNN0}. 
\end{proof}

%%-------------------- End of THEOREM ----------------------

Theorem~\ref{theorem:stabilityAlgNN0} highlights that the pointwise nonlinearities $\eta_\ell$ in the Gphon-NNs do not contribute to increase the difference between $\mathbf{\Phi}\left(\boldsymbol{x},\{ \mathcal{F}_{\ell} \}_{\ell=1}^{L},\boldsymbol{T}_{W}\right)$ and $\mathbf{\Phi}\left(\boldsymbol{x}, \{ \mathcal{F}_{\ell} \}_{\ell=1}^{L},\{ \boldsymbol{T}_{W_{G_\ell}} \}_{\ell=1}^{L} \right)$. This means that, any difference between an induced Gphon-NN from a GNN and its ideal Gphon-NN version on $W(x,y)$ is due to the operation of pooling, and this is captured in the upper bound in~\eqref{eq_thm_stabilityAlgNN0_0} stated in terms of the cut norm between two induced graphons. 
%Additionally, the value of this bound can be calculated with graphs obtained from $W(x,y)$ using methods \textbf{M1} and \textbf{M2}.

%%-----------------------------------
%%---------- REMARK -----------------
%%-----------------------------------

\begin{remark}\normalfont

Notice that the condition of being Lipschitz for the filters in Theorems~\ref{thm_change_filters} and \ref{theorem:stabilityAlgNN0} is analog to the condition of being low pass for filters in discrete signal processing (DSP). While the high frequencies in DSP are associated to the largest eigenvalues, high frequencies for graphon signals are associated to the lowest eigenvalues since zero is an accumulation point of the graphon shift operator. In fact, due to the nature of the spectrum of the graphon shift operator, even if the graphon filter $h(t)$ is Lipschitz it must be flatten out towards $t=0$.

\end{remark}

%%------- End of Remark -------------

Now we turn our attention to error estimates when we start with a graph approximation of an underlying graphon -- obtained by unspecified means -- and we perform the operation of pooling on those graph estimates.

%%-------------------------------------------
%%---------------- THEOREM ------------------
%%-------------------------------------------

\begin{theorem}
\label{thm_estimate_graphons}

Let $W(x,y)$ be a graphon and let $G_1$ and $G_2$ be two graph estimates of $W$ such that
$
\Vert
     \boldsymbol{T}_{W} - \boldsymbol{T}_{W_{G_i}}
\Vert_{HS} 
     \leq 
     \epsilon
     .
$
Let $H_i$ be the graph obtained from $W_{G_i}$ by method \textbf{M1}. If $\Vert \boldsymbol{T}_{W_{G_i}} - \boldsymbol{T}_{W_{H_i}} \Vert_{\infty\rightarrow 1}\leq \epsilon/V(H_i)^4$ we have that 

\begin{equation}
     \left\Vert 
            \boldsymbol{T}_{W_{H_1}} - \boldsymbol{T}_{W_{H_2}^{\theta}}
     \right\Vert_{\infty\rightarrow 1}
     \leq 32\epsilon
     ,
\end{equation}
where $W_{H_2}^{\theta} = W_{H_2}(\theta(x),\theta(y))$ and $\theta$ is any measure preserving map on $[0,1]$.
\end{theorem}

\begin{proof}
See Appendix~\ref{sec_proof_thm_estimate_graphons}.
\end{proof}

%%------------- End of THEOREM ----------------

Notice that the graphs $G_1$ and $G_2$ are discrete approximations of $W(x,y)$, obtained by arbitrary means. Then, Theorem~\ref{thm_estimate_graphons} shows that when the estimates $G_1$ and $G_2$ of $W(x,y)$ lead to $\boldsymbol{T}_{W}$ and $\boldsymbol{T}_{W_{G_i}}$ that are $\epsilon$-close in the Hilbert-Schmidt norm, we can apply the pooling method \textbf{M1} on $W_{G_i}$ to obtain graphs $H_i$ such that $\boldsymbol{T}_{W_{H_1}}$ and $\boldsymbol{T}_{W_{H_2}}$ are also $\epsilon$-close. This indicates that if two estimates of a graphon are close, the pooling method \textbf{M1} preserves this closeness. This fact endows the pooling method \textbf{M1} with a stability that is not guaranteed for the pooling approaches \textbf{M2} and \textbf{M3}, which also provides the theoretical support for its better performance shown in numerical simulations.

%%%%%%%%%%%%%%%%%%%%%%%%%%%%%%%%%%%%%%%%%%%
%%%%%%%%%%%%% SUB - SECTION %%%%%%%%%%%%%%%
%%%%%%%%%%%%%%%%%%%%%%%%%%%%%%%%%%%%%%%%%%%

\subsection{Label Permuation Equivariance and Invariance Attributes}

Although the pooling methods we propose are not label equivariant or invariant, we notice that Theorem~\ref{thm_estimate_graphons} has fundamental implications regarding the node labeling of the graphs when the graphon pooling method \textbf{M1} is applied. To see this, we note that one can consider three isomorphic graphs $G$, $G_1$ and $G_2$ -- i.e. the underlying graph is the same but the labeling of the nodes is different. Then, the term 
$
\Vert
     \boldsymbol{T}_{W_{G}} - \boldsymbol{T}_{W_{G_i}}
\Vert_{HS} 
     \leq 
     \epsilon     
$
provides a measure for the discrepancies between the node labelings associated to $G$, $G_1$ and $G_2$. We can think about the node labelings in $G$ as a baseline for comparison with the node labelings in $G_1$ and $G_2$, while the term 
$
\Vert \boldsymbol{T}_{W_{G_i}} - \boldsymbol{T}_{W_{H_i}} \Vert_{\infty\rightarrow 1}\leq \epsilon/V(H_i)^4
$
provides a measure of the changes produced in $G_i$ by the application of \textbf{M1}. Then, the term
$
\left\Vert 
            \boldsymbol{T}_{W_{H_1}} - \boldsymbol{T}_{W_{H_2}^{\theta}}
\right\Vert_{\infty\rightarrow 1}
  \leq 32\epsilon
$
indicates that $H_1$ and $H_2$ are close to be label permuation invariant under the lens of a graphon representation. In other words, Theorem~\ref{thm_estimate_graphons} implies that after applying \textbf{M1} we have a node label discrepancy that is proportional to the discrepancy in the node labelings of the original graphs where \textbf{M1} is being applied. If such discrepancy between $G_1$ and $G_2$ is $\epsilon$-small, the worst case scenario discrepancy between $H_1$ and $H_2$ is $32\epsilon$ under the lens of a graphon representation. This can be rephrased saying that \textbf{M1} is stable with respect to changes in the labels of the original graph, i.e. \textbf{M1} has quasi label permutation invariance/equivariance attributes.

%%%%%%%%%%%%%%%%%%%%%%%%%%%%%%%%%%%%%%%%%%%%%%%%%%%%
%%%%%%%%%%%%%%%% SUB - SECTION %%%%%%%%%%%%%%%%%%%%%
%%%%%%%%%%%%%%%%%%%%%%%%%%%%%%%%%%%%%%%%%%%%%%%%%%%%

\subsection{Edge Dropping Effects on Graphon Pooling}

In this section we discuss the effects of edge dropping on \textbf{M1}/\textbf{M2}/\textbf{M3}. Let us recall that for a graph $G$ with adjacency matrix $\bbA_{G}$, the process of edge dropping on $G$ produces a graph $\widehat{G}$ with adjacency matrix $\widehat{\bbA}_{\widehat{G}}$ given by
\begin{equation}\label{eq_edge_dropping_matrix}
 \widehat{\bbA}_{\widehat{G}} 
    =
       \bbA_{G}
       +
       \bbA_0
       ,
\end{equation}
where $\bbA_0$ is a perturbation symmetric matrix that cancels out some of the entries in $\bbA_{G}$. We can naturally extend~\eqref{eq_edge_dropping_matrix} to graphon representations as
\begin{equation}
 \widehat{\boldsymbol{T}}_{W_{\widehat{G}}}
      =
      \boldsymbol{T}_{W_G}
      +
      \boldsymbol{T}_0
      ,
\end{equation}
where $ \widehat{\boldsymbol{T}}_{W_{\widehat{G}}} $, $\boldsymbol{T}_{W_G}$ and $\boldsymbol{T}_0$ are the graphon shift operators associated with $\widehat{G}$, $G$ and $\bbA_0$, respectively. Taking into account this, we quantify the effects of edge dropping on \textbf{M1}/\textbf{M2}/\textbf{M3} in the following theorem.

%%-------------------------------------
%%----------- THEOREM -----------------
%%-------------------------------------

\begin{theorem}\label{thm_edge_drooping}

Let $W(x,y)$ be a graphon and let $G$ be the graph obtained from $W$ by means of \textbf{M1}/\textbf{M2}/\textbf{M3}. Let $\widehat{G}$ be the graph obtained from $G$ by edge dropping with graphon shift operator given by $\widehat{\boldsymbol{T}}_{W_{\widehat{G}}} = \boldsymbol{T}_{W_G} + \boldsymbol{T}_{0}$. Then, it follows that
\begin{equation}\label{eq_thm_edge_dropping_1}
  \left\Vert 
          \boldsymbol{T}_{W}
          -
          \widehat{\boldsymbol{T}}_{W_{\widehat{G}}}
 \right\Vert_2  
        \leq
 \left\Vert 
          \boldsymbol{T}_{W}
          -
          \boldsymbol{T}_{W_{G}}
 \right\Vert_2
        +
\left\Vert 
        \boldsymbol{T}_0
\right\Vert_2
.
\end{equation}
Additionally, if $h(t)$ is a $C$-Lipschitz filter we have
\begin{multline}\label{eq_thm_edge_dropping_2}
  \left\Vert 
      h\left(
          \boldsymbol{T}_{W}
        \right)
      -
      h\left(
          \widehat{\boldsymbol{T}}_{W_{G}}
        \right)
 \right\Vert_2  
        \leq
            \Omega \left( \bbT_{W} \right)
        +
        C(1+\delta)
        \left\Vert 
              \boldsymbol{T}_{0}
        \right\Vert_2
        \\
        +
        \ccalO\left( \Vert
                                  \boldsymbol{T}_{W}-\boldsymbol{T}_{W_{G}}
                               \Vert_{2}^2 
        \right)
        ,
\end{multline}
where $\Omega \left( \bbT_{W} \right)$ is specified according to Theorem~\ref{thm_change_filters} and $\delta$ is a constant that provides a measure of the non commutativity between $\boldsymbol{T}_{W_{G}}$ and $\boldsymbol{T}_{0}$, with $\delta=0$ if $\boldsymbol{T}_{W_{G}}$ and $\boldsymbol{T}_{0}$ commute.

\end{theorem}

\begin{proof} See Appendix~\ref{proof_thm_edge_drooping} \end{proof}

%%--------- End of Theorem ------------

The consequences of Theorem~\ref{thm_edge_drooping} are twofold. First, from~\eqref{eq_thm_edge_dropping_1} we can see that the results of Theorem~\ref{thm_M1andM2_convergence} will hold up to an additive factor of $\Vert \boldsymbol{T}_0 \Vert_2$. Then, how far is~\eqref{eq_thm_edge_dropping_1} from~\eqref{eq_thm_M1andM2_convergence} is dictated by the size of $\Vert \boldsymbol{T}_0\Vert_2$, i.e. the significance of the edges dropped. In practice this implies that one can consider edge dropping when implementing \textbf{M1}/\textbf{M2}/\textbf{M3} and in doing so Theorem~\ref{thm_M1andM2_convergence} will hold approximately as long as the edge dropped are negligible. Second, and similarly to the first scenario, we can see that the results from Theorem~\ref{thm_change_filters} hold approximately up to a factor of $C(1+\delta)\Vert \boldsymbol{T}_0 \Vert_2$ when edge dropping is considered. However, in this second scenario the variability of the filters, measured the the Lipschitz constant $C$, affects the significance of the edge dropping when applying Theorem~\ref{thm_change_filters}. Then, if the variation of a filter, $h(t)$, is not too high, Theorem~\ref{thm_change_filters} holds approximately in the presence of edge dropping.

%%-------------------------------------
%%------------ REMARK -----------------
%%-------------------------------------

\begin{remark}\normalfont
	
In the following section we perform a set of numerical experiments where we evaluate graphon pooling in GNNs. We remark that although the results derived above are expressed in terms of Gphon-NNs, they are directly associated to those GNNs that induce the Gphon-NNs. As we have stressed before in Remark~\ref{rmk_gphon_from_gnn}  there is a one to one relationship between the GNN with mapping operator $\mathbf{\Phi}\left( \bbx, \{ \ccalF_\ell \}_{\ell=1}^{L}, \{ \bbS_\ell \}_{\ell=1}^{L}\right)$ and its induced Gphon-NN with mapping operator given by $\mathbf{\Phi}\left( \boldsymbol{x}, \{ \ccalF_\ell \}_{\ell=1}^{L}, \{ \boldsymbol{T}_{W_{G_\ell}}\}_{\ell=1}^{L}\right)$. Then, we evaluate the proposed  graphon pooling methods with GNNs where graphs are constructed with the discretization methods (\textbf{M1}, \textbf{M2} and \textbf{M3}). Together with the numerical validations in the following section, the effectiveness of the graphon pooling operation can be proved both theoretically and numerically. Notice that for the application of \textbf{M1}/\textbf{M2}/\textbf{M3} we start using the induced graphon associated to an adjacency matrix $\mathbf{A}$, which is given by $W_{G}(x,y) = \mathbf{A}\left( 
\lceil Nx\rceil, \lceil Ny\rceil
\right)$ and no scaling on the adjacency matrix considered once we proceed to apply \textbf{M1}/\textbf{M2}/\textbf{M3}. Additionally, as mentioned in Remark~\ref{rmk_gphon_from_gnn}, we use the equivalence between GNNs and Gphon-NNs as an analytical tool to describe and understand the properties of the graphon pooling methods proposed, while numerical computations are performed on graphs and GNNs. This obeys to the fact that the scaling that determines such unique equivalence might affect severely numerical computations since a normalization of the graph shift operator by the number of nodes is not suitable when the graphs are sparse.

\end{remark}

%%----------- End of Remark -------------

%%%%%%%%%%%%%%%%%%%%%%%%%%%%%%%%%%%%%%%%%%%%%%%
%%%%%%%%%%%%%%% SUB - SECTION %%%%%%%%%%%%%%%%%
%%%%%%%%%%%%%%%%%%%%%%%%%%%%%%%%%%%%%%%%%%%%%%%

\subsection{Related work}
\label{sub_sec_related_work}

Graphons have been used previously for the general understanding of how information behaves on large size graphs. In~\cite{Diao2016ModelfreeCO} the first foundations of a signal processing theory are stated and used to study the problem of spectral clustering. In fact, the notion of node level statistic in~\cite{Diao2016ModelfreeCO} is what today is called a graphon signal, while the traditional spectral decomposition of the graphon operator acting on a signal is known today as the graphon Fourier transform. In~\cite{Morency2020GraphonFG,ruiz2020graphon} these results are formally stated in the language of graph signal processing and are used for the understanding of information processing models on graphs where the underlying graph is large. In~\cite{Ruiz2020GraphonNN} graphons are used as the building blocks of neural networks that at the same time are used to study the transferability properties of graph neural networks. None of the works mentioned considered the operation of pooling and in~\cite{Ruiz2020GraphonNN} such operation is not included in the definition and analysis of the so called graphon neural networks. We also remark that our error bounds and analytic guarantees rely more on the theory of Szemeredi partitions in the graphon space while in previous literature the results derived rely on an asymptotic analysis.

%% file: v16/sec_numsim.tex
%!TEX root =../compiler_paper.tex

%%%%%%%%%%%%%%%%%%%%%%%%%%%%%%%%%%%%%%%%%%%%%%%%
%%%%%%%%%%%%%%%%% SECTION %%%%%%%%%%%%%%%%%%%%%%
%%%%%%%%%%%%%%%%%%%%%%%%%%%%%%%%%%%%%%%%%%%%%%%%

\section{Numerical Simulations}\label{sec_numsim}

We consider three problem settings to verify the performance of our proposed graphon pooling methods, i.e. \textbf{M1}(regular integration), \textbf{M2}(irregular integration) and \textbf{M3}(irregular sampling). %We also consider a regular sampling method which is similar to \textbf{M3} but with a uniform grid as sampling points. 
By comparing with graph coarsening \cite{Defferrard2016ConvolutionalNN} and selection GNNs with zero padding \cite{gamagnns}, we claim that graphon pooling can achieve better performance while spending less time. All of the architectures are trained in parallel implementing the ADAM algorithm for stochastic optimization~\cite{kingma2014adam} with decaying factors set as $\beta_1= 0.9$ and $\beta_2 = 0.999$.

\subsection{Source localization problem}
\label{subsec:sourceloc}
We synthetically generate diffusion processes on graphs acquired from several graphon functions. The graph has $N$ nodes with the shift operator $\bbS \in \reals^{N \times N}$ and let $\bbx_0 \in \reals^N$ be a graph signal such that $[\bbx_0]_i = 1$ for node $i = c$ and $0$ otherwise. The graph diffusion process $\{\bbx_t\}$ is defined as 
\begin{equation}
\bbx_t = \bbS^t \bbx_0
\end{equation}
which represents the $t$-step diffusion graph signal. The objective is to locate the source node $c$ given $\bbx_t$ for arbitrary $t$. The source node $c$ is selected among $C=10$ possible sources, which makes this source localization problem as a classification problem on $C=10$ classes.

We train all the architectures with different pooling strategies to solve this problem by minimizing the cross-entropy loss on $1,000 \ (\bbx_t,c)$ training samples with learning rate $5 \times 10^{-4}$. The training dataset is divided in batches of $20$, over $300$ epochs. The learned architectures are validated and tested by evaluating the classification error rates on sets with $240$ and $200$ samples respectively. All the architectures are made up of GNNs with $L=2$ layers. Each layer contains $F_1 = F_2 = 8$ output features and $K_1 =K_2 =5$ filter taps.

We first analyze the performances of different graphon pooling methods on graphs obtained from 5 graphon models: the exponential graphon $W(x,y) = \exp (-\beta(x-y)^2)$ with $\beta=2.3$; the bilinear graphon $W(x,y) = xy$; and the polynomial graphon $W(x,y) = 0.5(x^2+y^2)$ together with a logarithmic graphon $W(x,y) = \log(1+\max(x,y))$ and an absolute graphon $W(x,y) =  |x-y|$ imported from \cite{xu2021learning}. The initial graphs with $N=100$ nodes are generated from function $W(x, y)$ by integration or sampling either regularly or irregularly. The number of selected nodes in the first and second layers in all architectures are $N_1 = 50$ and $N_2=25$ respectively. The layer-wise dimensionality reduction ratios are fixed as $2$ because of the setting of graph coarsening algorithm. The test classification error rates achieved with different graphon pooling methods, graph coarsening and the selection GNN are presented in Table  \ref{tb:graphons-source} for exponential, bilinear, polynomial, logarithmic and absolute graphons. We report the average error rate and standard deviation for models trained on $8$ different dataset realizations. We can observe from the results that graphon pooling and graph coarsening outperform selection GNN under an exponential, a logarithmic and an absolute graphon functions. Notice that \textbf{M2} and \textbf{M3} may lead to scenarios with some large sampling and integration intervals, which may cause those methods to not capture properly the behavior of the graphon in a portion of $[0,1]$. This is also consistent with the fact that for \textbf{M2} we do not have the error bound in~(\ref{eq_thm_change_filters_3}), since the partition grids associated to multiple applications of \textbf{M2} are not refined versions of each other, while (\ref{eq_thm_change_filters_3}) does not apply for \textbf{M3}. Furthermore, \textbf{M1} method can match or outperform graph coarsening method under all types of graphon functions. Though the graph coarsening method can achieve a stable performance, it still has limitations in calculation complexity and fixed dimensionality ratio.

%%-----------------------------------
%%------------ TABLE ----------------
%%-----------------------------------

\begin{table}%[t]
\centering
\subcaptionbox{Source localization with graphs initiated with \textbf{M1} graphon pooling \label{tb:graphons-ri}}
{\begin{tabular}{D|EEEEE} \hline
 					& \multicolumn{5}{c}{$W(x,y)$} \\
\makecell{Archi-\\tecture  }   		& $e^{-\beta(x-y)^2}$  & $xy$ & $0.5(x^2 + y^2)$  & {$\log(1+ \max(x,y))$}  & $|x-y|$   \\ \hline\hline 
\makecell{\textbf{M1}\\	pooling	}	& \makecell{$10.88 \pm $\\ $2.52$} & \makecell{$7.00 \pm $\\ $4.47$} & \makecell{$7.25 \pm $\\ $5.55$} & \makecell{$48.81 \pm $\\ $20.66$} & \makecell{$14.56 \pm $\\ $10.39$}  \\ \hline 
\makecell{Graph\\ coarsening}	& \makecell{$15.69 \pm $\\ $4.78 $} & \makecell{$6.56 \pm$\\ $ 3.55$} & \makecell{$7.10 \pm $\\ $4.39$}& \makecell{$35.88 \pm $\\ $17.76$}& \makecell{$27.44 \pm $\\ $19.08$} \\ \hline 
\makecell{Selection\\ GNN	}	& \makecell{$21.69 \pm $\\ $6.95$} & \makecell{$26.12 \pm $\\ $12.61$} & \makecell{$8.05 \pm $\\ $5.58$} & \makecell{$60.75 \pm $\\ $22.01$} & \makecell{$25.06 \pm $\\ $11.28$}  \\ \hline
\end{tabular}}
% \caption{Source localization test error rates (\%) achieved by regular integration graphon pooling, graph coarsening and selection GNN on $100$-node graphs obtained from exponential, bilinear and polynomial graphons.}
% \label{tb:graphons-ri}

% \vspace{2mm}
% \subcaptionbox{Source localization with graphs initiated with Regular Sampling (RS) graphon pooling. \label{tb:graphons-rs}}
% {
% \begin{tabular}{D|EEEEE} \hline
%  					& \multicolumn{5}{c}{$W(x,y)$} \\
% \makecell{Archi-\\tecture  }   		& $e^{-\beta(x-y)^2}$  & $xy$ & $0.5(x^2 + y^2)$  & {$\log(1+ \max(x,y))$}  & $|x-y|$   \\ \hline\hline
% \makecell{\textbf{M3}/RS	\\pooling}		& \makecell{$6.69 \pm $\\ $ 6.07$} & \makecell{$8.43 \pm $\\ $3.47$} & \makecell{$4.70 \pm$\\ $ 4.53$} & \makecell{$30.31 \pm$\\ $ 10.93$}& \makecell{$13.50 \pm$\\ $ 6.57$} \\ \hline
% \makecell{Graph \\ coarsening}	& \makecell{$6.13 \pm$\\ $ 5.01$} & \makecell{$7.97 \pm$\\ $ 4.25$} & \makecell{$9.30 \pm$\\ $ 4.06$} & \makecell{$21.21 \pm$\\ $ 10.48$} & \makecell{$15.69 \pm$\\ $ 4.39$}  \\\hline
% \makecell{Selection\\ GNN}		& \makecell{$17.12 \pm$\\ $ 12.67$} & \makecell{$13.07 \pm $\\ $9.64$} & \makecell{$13.10 \pm$\\ $ 7.40$} & \makecell{$56.56 \pm$\\ $ 18.49$} & \makecell{$18.75 \pm$\\ $ 7.55$}  \\ \hline
% \end{tabular}}

% \caption{Source localization test error rate (\%) achieved by graphon pooling, graph coarsening and selection GNN on $100$-node graphs obtained from exponential, bilinear and polynomial graphons.}
% \label{tb:graphons-rs}

\vspace{2mm}
\subcaptionbox{Source localization with graphs initiated with \textbf{M2} graphon pooling.\label{tb:graphons-ii}}
{
\begin{tabular}{D|EEEEE}\hline
 					& \multicolumn{5}{c}{$W(x,y)$} \\
\makecell{Archi-\\tecture  } 		& $e^{-\beta(x-y)^2}$  & $xy$ & $0.5(x^2 + y^2)$  & {$\log(1+ \max(x,y))$}  & $|x-y|$ \\ \hline \hline
\makecell{\textbf{M2} \\pooling	}	& \makecell{$21.79 \pm$\\ $11.18$} & \makecell{$10.75 \pm $\\ $7.43$} & \makecell{$23.30 \pm $\\ $2.50$} & \makecell{$ 55.25\pm$\\ $21.86 $} & \makecell{$ 13.81\pm$\\ $ 7.35$}\\  \hline
\makecell{Graph \\ coarsening}	& \makecell{$7.81 \pm$\\ $ 4.49$} & \makecell{$4.69 \pm$\\ $ 3.69$} & \makecell{$6.40 \pm$\\ $ 5.60$} & \makecell{$ 38.44\pm$\\ $ 12.96$} &\makecell{$ 18.31\pm$\\ $ 10.62$} \\ \hline
\makecell{Selection \\ GNN}		& \makecell{$36.75 \pm$\\ $ 11.87$} & \makecell{$8.62 \pm$\\ $ 6.61$} & \makecell{$26.90 \pm$\\ $ 6.51$} &\makecell{ $ 67.19 \pm$\\ $ 21.83$} &\makecell{$ 19.75\pm$\\ $ 7.03$} \\ \hline
\end{tabular}
}

\vspace{2mm}
\subcaptionbox{Source localization with graphs initiated with \textbf{M3} graphon pooling.\label{tb:graphons-ii}}
{
\begin{tabular}{D|EEEEE} \hline
 					& \multicolumn{5}{c}{$W(x,y)$} \\
\makecell{Archi-\\tecture  }   		& $e^{-\beta(x-y)^2})$  & $xy$ & $0.5(x^2 + y^2)$ & {$\log(1+ \max(x,y))$}  & $|x-y|$  \\ \hline \hline
\makecell{\textbf{M3} \\	pooling	}	& \makecell{$19.81 \pm$\\ $ 11.02$} & \makecell{$11.50 \pm $\\ $8.69$} & \makecell{$17.69 \pm $\\ $7.17$}  & \makecell{$52.69 \pm $\\ $23.58$} & \makecell{$23.56 \pm $\\ $5.49$}\\  \hline
\makecell{Graph\\ coarsening}	& \makecell{$9.31 \pm $\\ $7.87$} & \makecell{$9.83 \pm $\\ $9.74$} & \makecell{$12.31 \pm $\\ $7.02$} & \makecell{$23.19 \pm $\\ $17.42$}& \makecell{$29.31 \pm $\\ $13.74$} \\ \hline
\makecell{Selection\\ GNN	}	& \makecell{$34.50 \pm$\\ $18.20$} & \makecell{$11.33 \pm $\\ $8.73$} & \makecell{$16.56 \pm $\\ $7.78$} & \makecell{$57.50 \pm $\\ $18.44$} & \makecell{$38.12 \pm $\\ $10.50$}  \\ \hline
\end{tabular}
}

\caption{Source localization test error rates (\%) achieved by four graphon pooling methods, graph coarsening and selection GNN on $100$-node graphs obtained from exponential, bilinear and polynomial graphons.}
\label{tb:graphons-source}
\end{table}

We further investigate the influence of  the layer-wise dimensionality reduction ratios $N_1/N$ (layer 1) and $N_2/N_1$ (layer 2) on the performance. We focus on the polynomial graphon $W(x,y) = 0.5(x^2+y^2)$ with different number of nodes $N$ and the numbers of selected nodes $N_1$ and $N_2$ as shown in the rows of Table \ref{tb:ratios}. Considering the fixed sampling ratios in graph coarsening, we only compare {\textbf{M1}} graphon pooling method with selection GNN. This also shows that our graphon pooling also perform well with reduced complexity, especially when the reduction ratio is large.

\begin{table}%[t]
%\scriptsize
\centering
\subcaptionbox{Source localization with graphs initiated with \textbf{M1} graphon pooling.\label{tb:graphons-ri-ratios}}
{
\begin{tabular}{c|c|cc} \hline
$[N,N_1,N_2]$   	& $\left[\dfrac{N}{N_1},\dfrac{N_1}{N_2}\right]$  & \textbf{M1} pooling  & Selection GNN \\ \hline
$[100,50,10]$		& $[2,5]$ & $5.88 \pm 4.71$    & $4.56\pm 3.9$ \\
$[200,100,10]$		& $[2,10]$ & $23.40 \pm 12.88$    & $25.70 \pm 8.33$ \\ 
$[400,200,10]$		& $[2,20]$ & $28.94 \pm 9.63$   & $38.81 \pm 16.41$ \\ \hline
$[100,20,10]$		& $[5,2]$ & $10.31\pm 7.36$    & $21.25 \pm 10.79$ \\ 
$[200,20,10]$		& $[10,2]$ & $29.56 \pm 12.74$    & $44.19 \pm 14.74$ \\  
$[400,20,10]$		& $[20,2]$ & $46.05 \pm 14.95$   & $54.70 \pm 12.58$ \\  \hline
\end{tabular}
}

\caption{Source localization test error rates (\%) achieved by {\textbf{M1}} graphon pooling, graph coarsening and selection GNN on graphs obtained from $W(x,y) = 0.5(x^2 + y^2)$ with different values of $N$, $N_1$ and $N_2$.}
\label{tb:ratios}
\end{table} 

\subsection{Point cloud classification problem}
\label{subsec:pointcloud}
We next evaluate the pooling strategies on the ModelNet10~\cite{wu20153d} to classify a certain object. The ModelNet10 dataset contains 3,991 meshed CAD models from 10 categories for training and 908 models for testing. We sample $300$ points from each model to construct a point cloud. Each point possesses a 3D coordinate as features. We model the graph by seeing the sampling points as nodes and the distance between every pair of nodes as edge weight between two nodes. Based on this graph, we can generate a step graphon function with $300 \times 300$ blocks. Our goal is to identify the models for chairs from all the other categories.

%%-------------------------------------
%%------------ FIGURE ----------------- 
%%-------------------------------------

\begin{figure}[t]
\centering
			 
\includegraphics[width=0.12\textwidth]{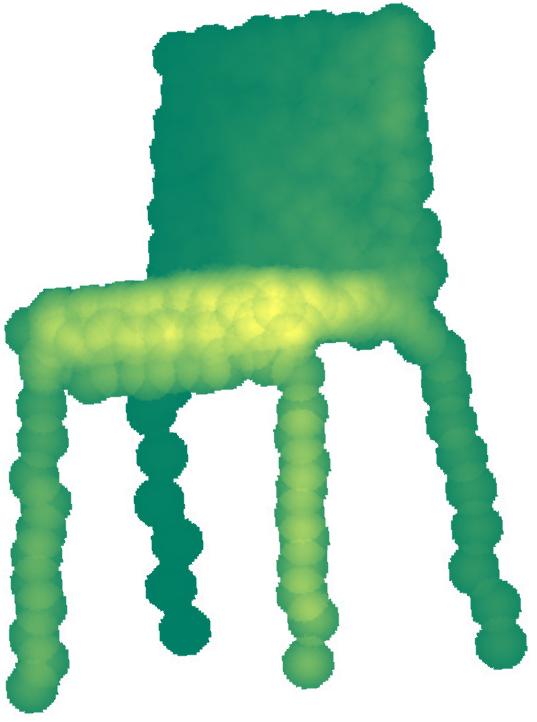}\hspace{4mm}
\includegraphics[width=0.12\textwidth]{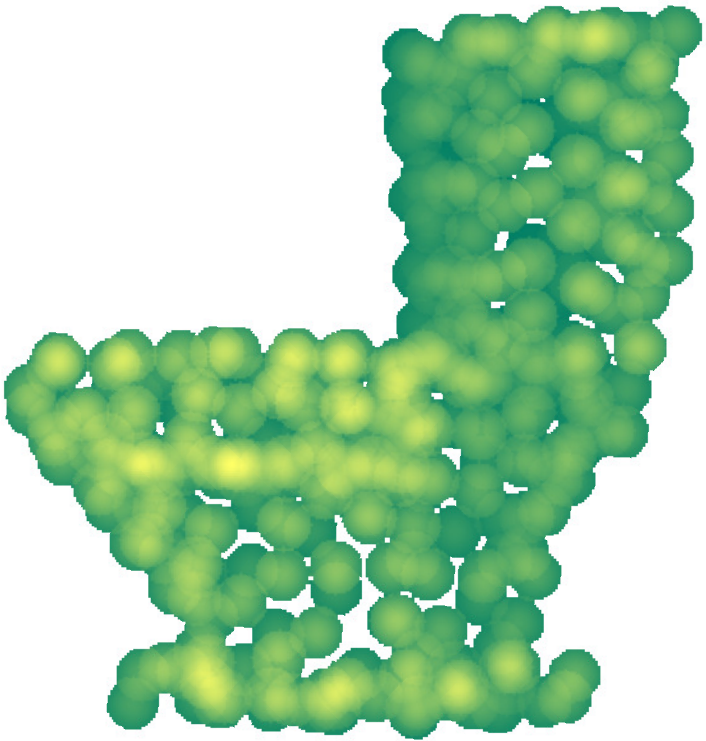} \hspace{4mm}
\includegraphics[width=0.12\textwidth]{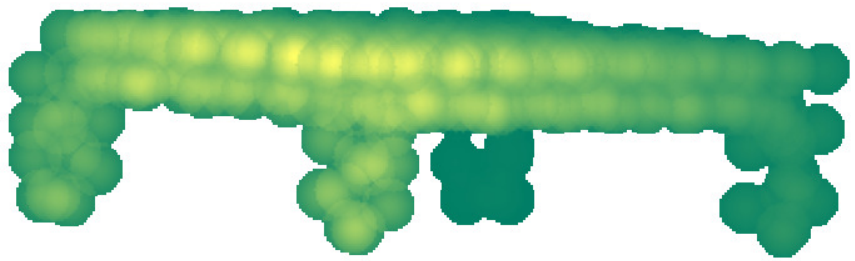} 

\caption{Point cloud models with 300 sampling points in each model. Our goal is to identify chair models from other models such as toilet and table. }
\label{fig:modelnet}
\end{figure}

%%--------- End of Figure -------------

We implement graphon pooling, graph coarsening, selection GNN and an adaptive DiffPool \cite{ying2018hierarchical} architectures. All these architectures include GNNs with 2 layers with $F_1=64$ output features in the first layer and $F_2= 32$ output features in the second layer. Each layer contains $K=5$ filter taps. We use ReLU as the activation function. All architectures also include a final readout layer to map the graph output features to a binary classfication scalar. For the DiffPool architecture, we adapt the architecture proposed in \cite{ying2018hierarchical} and construct two pooling architectures following each graph filter layer. The pooling architecture includes a graph filter layer, a nonlinearity and a linear readout layer. All the architectures are trained by minimizing the cross-entropy loss with the learning rate set as $0.005$. We divide the training models into batches of $10$ models over $40$ epochs. We repeat $5$ sampling realizations for all the architectures and evaluate the performance by averaging the classification error rates as well as the standard deviation in Table \ref{tb:pointcloud-err}.

We can observe that \textbf{M1} graphon pooling method outperforms graph coarsening and selection GNN when the dimensionality ratio is large as the last two columns show in Figure \ref{tb:pointcloud-err}. We also observe that \textbf{M1} graphon pooling nearly matches the parametric DiffPool method where an adaptive pooling strategy is learned for each layer. We further compare the computation complexity by demonstrating the training time per batch spent in each strategy as Table \ref{tb:pointcloud-time} shows. We claim that though graphon pooling may achieve similar results or slightly worse results when compared with graph coarsening and DiffPool, it needs much less time for training. This indicates the graphon pooling method can achieve a compatible performance with high computation efficiency.

\begin{table}[t]
\centering
\begin{tabular}{A|CCC}
\hline
 					& \multicolumn{3}{c}{$[N_1,N_2]$} \\

Architecture  & $ [150, 75]$   & $  [100,50]$  & $  [50,10]$  \\ \hline
\textbf{M1} 	pooling	& $2.64  \pm 0.27 $& $3.86  \pm 0.76 $ & $4.99 \pm 0.73$  \\ \hline
% RS  pooling & $2.72 \pm 0.44$ 		& $4.60 \pm 0.43$ & $5.00 \pm 0.73 $  \\ \hline
\textbf{M2}  pooling	& $3.45 \pm 0.11$ 	& $4.76 \pm 0.58$ & $5.36 \pm 0.81 $  \\ \hline
\textbf{M3}  pooling	& $3.38 \pm 0.60$	& $4.94 \pm 0.42$ & $7.01 \pm 2.93 $  \\ \hline
Selection GNN		& $2.25\pm 0.46$	& $4.79\pm 0.80$ & $6.01 \pm 0.67 $  \\ \hline
Graph Coarsening		& $2.01\pm 0.36$	& -- & -- \\ \hline
Differential pooling & $2.37\pm 0.56$	& $4.02\pm 1.23$ & $4.89\pm 1.32$ \\ \hline
\end{tabular}
\caption{Prediction error rates (\%) for model `chair' in the test dataset. Average over 5 data realizations. The number of nodes is $N=300$, and $[N_1,N_2]$ stands for the number of selected nodes in the first and second layers of the GNN.}
\label{tb:pointcloud-err}
\end{table} 

\begin{table}[t]
\centering
\begin{tabular}{l|c|c} \hline
Architecture    & $[N_1,N_2] = [150,75]$  & $ [N_1,N_2] = [50,10]$  \\ \hline
\textbf{M1}  pooling	& $0.401s \pm 0.012s $ & $0.070s \pm 0.008s$  \\ \hline
% RS  	pooling		& $0.079s \pm 0.006s$ & $0.026s \pm 0.008s $  \\ \hline
\textbf{M2}  pooling	& $0.419s \pm 0.012s$ & $0.076s \pm 0.008s $  \\ \hline
\textbf{M3}  pooling	& $0.081s \pm 0.006s$ & $0.027s\pm 0.008s $  \\ \hline
Selection GNN		& $1.861s \pm 0.089s$ & $ 1.089s \pm 0.028s $  \\ \hline
Graph coarsening		& $ 5.283s\pm 0.056s $ &  -- \\ \hline
Differential pooling & $3.982s \pm 0.322s$ & $ 3.762s \pm 0.317s $  \\ \hline
\end{tabular}
\caption{Average training time spent per batch in point cloud classification problem. The number of nodes is $N=300$, and $[N_1,N_2]$ stands for the number of selected nodes in the first and second layers of the GNN.}
\label{tb:pointcloud-time}
\end{table}

\subsection{Recommendation system problem}
We implement the MovieLens 100k dataset~\cite{harper2015movielens} to construct a user similarity network. The MovieLens dataset contains 100,000 ratings given from 943 users to 1,682 movies. By calculating Pearson correlations between the ratings given by two users to the same movies~\cite{huang2018rating} while keeping the number of nearest neighbors of each user as 50, we can build a full similarity network. A step graphon function with $943 \times 943$ blocks can be generated based on this full network. 

The graph signal represents the movies' rating vectors with the $u$-th element of the rating vector for movie $m$ standing for the rating given by user $u$ to movie $m$. Given an incomplete rating vector of a specific movie, we can predict a user's rating based on the similarity network.

We train GNNs with graphon pooling and selection GNN architectures by minimizing the mean squared error (MSE) loss between the real and the predicted ratings. All architectures contain 2 layers with $F_1 = 32$ and $F_2 = 8$ features respectively. Each layer consists of $K_1=K_2=5$ filter taps. We focus on user 1 and divide 90\% of movie ratings from user 1 for training while the rest for testing. We train all the architectures over $40$ epochs with batch size $5$.

We set the number of selected nodes as $N_1 = 100$ and $N_2 = 10$ nodes in the first and second layers respectively. We also consider another setting with $N_1 = 50$ and $N_2 = 10$. The average prediction RMSEs and the standard deviations over 10 different data realizations are presented in Table \ref{tb:movies}. For $[N_1, N_2] = [100,10]$, the GNN with graphon pooling achieves lower test RMSE than the selection GNNs especially the \textbf{M1} method, which is accordant with our conclusions in Section \ref{subsec:sourceloc} and \ref{subsec:pointcloud}.

\begin{table}[t]
\centering
\begin{tabular}{l|c|c} \hline
Architecture    & $[N_1,N_2] = [100, 10]$  & $ [N_1,N_2] = [50,10]$  \\ \hline
\makecell{\textbf{M1} 	pooling	}	& $1.1066\pm 0.1497$ & $1.1678 \pm 0.1158$  \\ \hline
% \makecell{RS	pooling}		& $1.1121 \pm 0.1462$ & $1.1797 \pm 0.1358$  \\ \hline
\makecell{\textbf{M2}   pooling}		& $1.1558 \pm 0.1342$ & $1.1998 \pm 0.1380$  \\ \hline
\makecell{\textbf{M3}   pooling}		& $1.1132 \pm 0.1471$ & $1.2160 \pm 0.1301$  \\ \hline
Selection GNN		& $1.1339 \pm 0.1152$ & $1.2167 \pm 0.1006$  \\ \hline
\end{tabular}
\caption{Prediction RMSE for user 1's ratings to movies in the test set. Average over 10 train-test splits. The number of nodes is $N=943$, and $[N_1,N_2]$ stands for the number of selected nodes in the first and second layers of the GNN.}
\label{tb:movies}
\vspace{-3mm}
\end{table}

%% file: v16/sec_discussion.tex
%!TEX root =../compiler_paper.tex

%%%%%%%%%%%%%%%%%%%%%%%%%%%%%%%%%%%%%%%%%%%%%%%%%%%%%%%%
%%%%%%%%%%%%%%%%%%% SECTION %%%%%%%%%%%%%%%%%%%%%%%%%%%%
%%%%%%%%%%%%%%%%%%%%%%%%%%%%%%%%%%%%%%%%%%%%%%%%%%%%%%%%

\section{Discussion and Conclusions}
\label{sec_discussion}

We proposed three pooling strategies for signals and operators on graphs based on graphon representations -- Section~\ref{sec_graphon_pooling}, methods \textbf{M1}, \textbf{M2}, and \textbf{M3}. The underlying idea of \textbf{M1}/\textbf{M2}/\textbf{M3} relies on building sequences of graphs and graph signals from graphon representations. We tested our approach on GNNs, making a comparison with other graph pooling approaches such as graph coarsening and zero padding -- Section~\ref{sec_graphon_pooling}.

We proved that methods \textbf{M1}/\textbf{M2} based on integration over partitions of $[0,1]^2$ lead to bounded errors for the filters and mapping operators when compared to a Gphon-NN built only with the original graph/graphon -- Theorems~\ref{thm_M1andM2_convergence},~\ref{thm_change_filters}, and~\ref{thm_estimate_graphons}. In Theorem~\ref{thm_M1andM2_convergence} we showed that the shift operators on the induced graphons of the graphs obtained after applying \textbf{M1}/\textbf{M2} are close to the shift operator on the induced graphon of the original graph. This closeness is inversely proportional to the squared number of elements of the partition considered. In Theorem~\ref{thm_change_filters} we showed that the size of the difference between the filters on the resultant induced graphon after applying \textbf{M1}/\textbf{M2} and the induced graphon of the original graph, is bounded by the squared root of the cut norm distance between graphon representations. This bound can be calculated using graphs/graphons obtained after applying \textbf{M1}/\textbf{M2}. Theorem~\ref{theorem:stabilityAlgNN0} extends Theorem~\ref{thm_change_filters} to the mapping operators of the Gphon-NNs, showing that pointwise nonlinearities do not alter the error associated to the pooling operation. Additionally, we formally showed that the method \textbf{M1} -- integration over an equipartition of $[0,1]$ -- is stable to arbitrary approximations of the graph and its induced graphon representation. This is, given two graph approximations of a graphon that are $\epsilon$-close, we have that \textbf{M1} preserves the closeness between such approximations up to a scalar factor -- Theorem~\ref{thm_estimate_graphons}. Notice that although \textbf{M3} provides the poorest performance among the pooling methods presented, it has the advantage of providing a low cost option to perform pooling that relies on graphon representations.

The numerical experiments in Section~\ref{sec_numsim} corroborate the results derived from our analysis in Section~\ref{sec_graphon_pooling}, showing that among the graphon pooling methods proposed \textbf{M1} provides the best results. This points to a unique attribute of equipartitions in the graphon space, which is partly explained by the stability of \textbf{M1} to arbitrary approximations of the graph to be reduced -- Theorem~\ref{thm_estimate_graphons}.

When applied to reduce dimensionality between layers in GNNs, graphon pooling methods adapt particularly well to those scenarios where the graph in the first layer is large. This is a consequence of two fundamental facts. First, graphons naturally model graphs of large size. Second, the graphon pooling methods are more effective than other pooling approaches when there is a large dimensionality reduction ratio between the original signal/operator and the reduced one.

Although valuable for medium size graphs, pooling methods such as graph coarsening and zero padding are not applicable for large graphs because of their computational cost. In contrast with this, graphon pooling adapts naturally to large graphs since graphons are by themselves limit objects of sequences of graphs whose number of nodes and edges grows up to infinity.

There is a rich research direction to explore the reduction of the graph signal involving directly the structure of the grid used in \textbf{M2}. In particular, considering different weighting measures on the signal considering the size of the partition associated to each subelement in the irregular grid.

We showed that \textbf{M1} is endowed with an approximate label permuation invariance/equivariance principle. Given the benefits exhibited by graphon pooling methods, we consider that an interesting future problem is that establishing concrete properties on the labeling of the graph under which label permutation invariance and equivariance can be achieved for \textbf{M2} and \textbf{M3}.

%% file: v16/sec_appendix.tex
%!TEX root =../compiler_paper.tex

%%%%%%%%%%%%%%%%%%%%%%%%%%%%%%%%%%%%%%%%%%%
%%%%%%%%%%%%%%% SECTION %%%%%%%%%%%%%%%%%%%
%%%%%%%%%%%%%%%%%%%%%%%%%%%%%%%%%%%%%%%%%%%

\section{Background material}
\label{sec_appendix_background}

%%%%%%%%%%%%%%%%%%%%%%%%%%%%%%%%%%%
%%%%%%%%%%%%% SUB-SECTION %%%%%%%%%
%%%%%%%%%%%%%%%%%%%%%%%%%%%%%%%%%%%

\subsection{Basic preliminaries}

%%%%%%%%%%%%%%%%%%%%%%%%%%%%%%%%%%%
%%%%%%%%%% Sub-Sub-Secction %%%%%%%
%%%%%%%%%%%%%%%%%%%%%%%%%%%%%%%%%%%

\subsubsection{Spectral representation of convolutional filtering on graphs}
If $\bbS$ is diagonalizable, we can leverage the spectral representation of $\bbS$ to define a Fourier transform and a spectral representation of the filters. Let $\bbS = \bbU \mathbf{\Lambda} \bbU^{\mathsf{T}}$, where $\mathbf{\Lambda}$ is diagonal and $\bbU$ is orthogonal. Then, we say that $\hat{\bbx} = \bbU^{\mathsf{T}}\bbx$ is the graph Fourier transform (GFT) of $\mathbf{x}$. The filtering of $\bbx$ can be expressed in the Fourier domain according to
%
%
%
%\begin{equation}\label{eq_Hingraph_spec}
$
\mathbf{H}(\mathbf{S})\mathbf{x}
                  =
                    \bbU
                    \sum_{k=0}^{K-1}h_{k}\mathbf{\Lambda}^{k}\hat{\mathbf{x}}
.
$
%\end{equation}
%
%
%
This is, to do the filtering of $\bbx$ by $\bbH(\bbS)$ we can do the spectral filtering of $\hat{\bbx}$ as 
$ 
\sum_{k=0}^{K-1}h_{k}\mathbf{\Lambda}^{k}\hat{\mathbf{x}}
,
$
and then take the inverse Fourier transform given by the right action of $\bbU$.

%%%%%%%%%%%%%%%%%%%%%%%%%%%%%%%%%%%
%%%%%%%%%% Sub-Sub-Secction %%%%%%%
%%%%%%%%%%%%%%%%%%%%%%%%%%%%%%%%%%%

\subsubsection{Spectral representation of convolutional filtering on graphons}

Using spectral decompositions of $\boldsymbol{T}_W$, it is possible to define a  Fourier transform on graphons. Suppose $\lambda_i (\boldsymbol{T}_W)$ and $\boldsymbol{\varphi}_{W,i}$ are the $i$-th eigenvalue and $i$-th eigenvector of $\boldsymbol{T}_W$, respectively. We denote the graphon Fourier transform (Gphon-FT) of $(W,\boldsymbol{x})$ by $(\widehat{W},\hat{\boldsymbol{x}})$, where $\hat{\boldsymbol{x}}\in \ell_2 \left( \mathbb{Z} \right)$ with

\begin{equation}
\hat{\boldsymbol{x}}(j)
              =
                  \int_{0}^{1}\boldsymbol{x}(u)\boldsymbol{\varphi}_{W,j}(u)du
,
\end{equation}
and where the symbol $\widehat{W}$ emphasizes that $\hat{\boldsymbol{x}}$ is defined on the spectrum -- eigenvalues -- of $\boldsymbol{T}_{W}$ and not in $[0,1]$.

There is a natural connection between the GFT and the Gphon-FT when a graphon signal is induced by a graph signal. If $W_G$ is induced by $G$, we have $\lambda_i(\boldsymbol{T}_{W_G}) = \lambda_i (G)/\vert V(G)\vert$ for $i=1,\ldots, \vert V(G)\vert$ and $\lambda_i(\boldsymbol{T}_{W_G}) = 0$ for all $i>\vert V(G)\vert$~\cite{lovaz2012large,Diao2016ModelfreeCO}, considering an ordering of the eigenvalues in decreasing absolute modulus. Furthermore, if $\bbu_i$ is the $i$-th eigenvector of $\bbS$ -- the shift operator of the graph $G$ --, the $i$-th eigenvector of $\boldsymbol{T}_{W_G}$ is given by $\boldsymbol{\varphi}_{W_G,i} (t) =\sqrt{\vert V(G)\vert}\bbu_i ( \lceil t\vert V(G)\vert \rceil )$ with $N = \vert V(G) \vert$~\cite{gao2019graphon}.

The convergence of a sequence of graphs to a graphon has implications on the convergence of spectral representations on the graphs and the graphons~\cite{Diao2016ModelfreeCO}. This is formally stated in the following theorem from~\cite{Diao2016ModelfreeCO}.

%%--------------------------------------------
%%---------------- THEOREM -------------------
%%--------------------------------------------

\begin{theorem}[5.6~\cite{Diao2016ModelfreeCO}]\label{thm_con_eigen}
	
	Let $\{ W_i \}_i$ be a sequence of graphons uniformly bounded in $L^{\infty}$. Suppose $\Vert W_{i} -W \Vert_{\square}\rightarrow 0$ as $i\rightarrow\infty$. For $k\geq 1$, let $P_{k}(W_i):L^{2}([0,1])\rightarrow L^{2}([0,1])$ be the projection operator on the eigenspace of $W_i$ associated to the eigenvalues $\{  \lambda_{k}(W_i) , - \lambda_{k}(W_i) \}$. For all $k\geq 1$ we have
	\begin{enumerate}
		\item $\displaystyle \lambda_{k}(W_i)\rightarrow \lambda_{k}(W)$ as $i\rightarrow\infty$,
		\item $\displaystyle\left\Vert P_{k}(W_i) - P_{k}(W) \right\Vert_{2} \rightarrow 0$ as $i\rightarrow\infty$.
	\end{enumerate}
\end{theorem}

%%-------------- End of Theorem -----------------

%%%%%%%%%%%%%%%%%%%%%%%%%%%%%%%%%%%
%%%%%%%%%% Sub-Sub-Secction %%%%%%%
%%%%%%%%%%%%%%%%%%%%%%%%%%%%%%%%%%%

\subsubsection{Convergence of homomorphism densities}
\label{subsec_basic_preliminaries}

In this subsection we discuss the notion of convergence of sequences of graphs to graphs, relying on the notion of homomorphisms density. A homomorphism from the graph $H$ to the graph $G$ is an edge preserving map from $V(H)$ to $V(G)$. If we denote by $\Hom (H,G)$ the number of homomorphisms between $H$ and $G$, it is possible to define a homomorphism density $t(H,G)$ by
%
%
%
%\begin{equation}
$
t(H,G)
        =
\left\vert\Hom(H,G)\right\vert
/
\vert V(G)\vert^{\vert V(H)\vert}
.
$
%\end{equation}
%
%
%
This notion of density can be extended to compare graphs and graphons. If $H$ is a graph and $W(x,y)$ is a graphon, the density of homomorphisms between $H$ and $W(x,y)$ can be calculated as
\begin{equation}
t(H , W)
= 
\int_{[0,1]^{\vert V(H)\vert}} 
\prod_{\{ i, j\} \in E(H) } W(x_i , x_j)dx_1 \ldots dx_{\vert V(H) \vert} 
.
\end{equation}
If $ W_G$ is a graphon induced by a graph $G$, then it is possible to show that $t(H,W_{G}) = t(H , G)$~\cite{Diao2016ModelfreeCO,lovaz2012large,Glasscock2015WhatIA}.

We say that a sequence of graphs $\{ G_i \}_{i=0}^{\infty}$ converge to the graphon $W(x,y)$ -- we denote this by $\{ G_i \}\rightarrow W(x,y)$ -- if $\lim_{i\rightarrow\infty}t(H,G_i) = t(H,W)$ for all simple finite graphs $H$.

%%%%%%%%%%%%%%%%%%%%%%%%%%%%%%%%%%%%%%%%%%%
%%%%%%%%%%%%%%% SECTION %%%%%%%%%%%%%%%%%%%
%%%%%%%%%%%%%%%%%%%%%%%%%%%%%%%%%%%%%%%%%%%

\section{Proofs}
\label{sec_appendix_proofs}

We start introducing some notation that will facilitate the presentation of the proofs. The symbol $\ccalB (\ccalV)$ indicates the set of bounded operators acting on $\ccalV$. The norm indicated as $\Vert \cdot \Vert_{2}$ represents the norm $\Vert \cdot \Vert_{2\rightarrow 2}$ on operators in $\ccalB (L^{2}([0,1]))$ -- the set of bounded operators acting on $L^{2}([0,1])$. The norm indicated as $\Vert \cdot \Vert_{op2}$ represents the norm $\Vert \cdot \Vert_{\ccalB (L^{2}([0,1]))\rightarrow \ccalB (L^{2}([0,1]))}$ acting on elements of $\ccalB \left( \ccalB \left( L^{2}([0,1]) \right) \right)$. The symbol $\Vert \cdot \Vert_{op(\ccalB\otimes\ccalB)}$ represents the operator norm of an operator living in $\ccalB (L^2 ([0,1]))^{\ast}\otimes\ccalB (L^2 ([0,1]))$. The symbol $\Vert \cdot \Vert_{L^2 \otimes L^2}$ is the norm acting on $L^2 ([0,1])\otimes L^2 ([0,1])$. Given the graphon $W(x,y)$ we will denote by $\Vert W \Vert_{1}$ and $\Vert W \Vert_{2}$ the $L^1$ and $L^2$ norms of $W(x,y)$ as elements of $L^{1} \left( [0,1]^{2}\right)$ and $L^{2} \left( [0,1]^{2} \right)$.

%%%%%%%%%%%%%%%%%%%%%%%%%%%%%%%%%%%
%%%%%%%%%%%%% SUB-SECTION %%%%%%%%%
%%%%%%%%%%%%%%%%%%%%%%%%%%%%%%%%%%%

\subsection{Proof of Theorem~\ref{thm_induced_gphon_filtering}}
\label{proof_thm_induced_gphon_filtering}

First, we start taking into account that by means of the spectral theorem we have
\begin{equation}
\boldsymbol{y}
    =
    h(\boldsymbol{T}_{W_G})\boldsymbol{x}
    =
    \sum_{i=1}^{\infty}
    h\left(
        \lambda_i(\boldsymbol{T}_{W_G})
    \right)
    \boldsymbol{\varphi}_{W_G,i}
    \left\langle
         \boldsymbol{\varphi}_{W_G,i}
         ,
         \boldsymbol{x}
    \right\rangle
    .
\end{equation}
Taking into account that
\begin{equation}
    \boldsymbol{\varphi}_{W_G,i}
    =
    \sqrt{\vert V(G)\vert}
     \sum_{r=1}^{\vert V(G)\vert}\bbu_i(r)\chi_{r}(u)   
    ,
    \quad
    \boldsymbol{x} 
           = \sum_{r=1}^{\vert V(G)\vert}\bbx(r)\chi_{r}(u)
           ,
\end{equation}
where $\bbu_i$ is the $i$-th eigenvector of $\bbS_G$ and $\chi_{r}(u)$ is the characteristic function of the interval $[(r-1)/\vert V(G)\vert, r/\vert V(G)\vert]$, it follows that
\begin{multline}
\left\langle
         \boldsymbol{\varphi}_{W_G,i}
         ,
         \boldsymbol{x}
\right\rangle
=
\\
\int_{0}^{1}
        \sqrt{\vert V(G)\vert}
     \sum_{r=1}^{\vert V(G)\vert}\bbu_i(r)\chi_{r}(u) 
    \sum_{r=1}^{N}\bbx(r)\chi_{r}(u)du
    \\
=  
      \frac{
       \left\langle
         \bbu_i
         ,
         \bbx
       \right\rangle     
      }
      {\sqrt{\vert V(G) \vert}}
      .
\end{multline}
Then, taking into account that $\lambda_i(\boldsymbol{T}_{W_G}) = \lambda_i (G)/\vert V(G)\vert$ and $ \boldsymbol{\varphi}_{W_G,i} = \sqrt{\vert V(G)\vert}\mathsf{step}(\bbu_i)$ we reach
\begin{equation}
\boldsymbol{y}
    =
    \sum_{i=1}^{\infty}
    h\left(
        \frac{\lambda_i (G)}{\vert V(G)\vert}
    \right)
    \sqrt{\vert V(G)\vert}\mathsf{step}(\bbu_i)
      \frac{
       \left\langle
         \bbu_i
         ,
         \bbx
       \right\rangle     
      }
      {\sqrt{\vert V(G) \vert}}
    .
\end{equation}
Using the properties of $\mathsf{step}(\cdot)$ we have
\begin{equation}
\boldsymbol{y}
    =
\mathsf{step}
    \left(
    \sum_{i=1}^{\infty}
    h\left(
        \frac{\lambda_i (G)}{\vert V(G)\vert}
    \right)
       \bbu_i
       \left\langle
         \bbu_i
         ,
         \bbx
       \right\rangle     
\right)
    ,
\end{equation}
and by means of the spectral theorem it follows that
\begin{equation}
\boldsymbol{y}
          =
          \mathsf{step}
          \left(
           h\left( 
                 \frac{\bbS_G}{\vert V(G)\vert}
           \right)\bbx
          \right) 
.
\end{equation}
%
%
%

%%%%%%%%%%%%%%%%%%%%%%%%%%%%%%%%%%%
%%%%%%%%%%%%% SUB-SECTION %%%%%%%%%
%%%%%%%%%%%%%%%%%%%%%%%%%%%%%%%%%%%

\subsection{Proof of Theorem~\ref{thm_M1andM2_convergence}}
\label{proof_thm_M1andM2_convergence}

\begin{proof}

This proof follows directly from the application of Proposition 9.8 in~\cite{lovaz2012large}. To see this, we start recalling the notation in~\cite{lovaz2012large} where for a given graphon $W(x,y)$ its \textit{stepping} is given by

\begin{equation}
W_{\ccalP}(x,y) 
            =
               \frac{1}{\mu(\ccalS_i)\mu(\ccalS_j)}
               \int_{\ccalS_i \times \ccalS_j}W(x,y)dxdy
               ,
\end{equation}
where $\mu$ is a measure function and $\ccalP = \{ \ccalS_i \}_i$ is a partition of $[0,1]$.  Now, we recall Proposition 9.8 in~\cite{lovaz2012large}.

%%-------------------------------------
%%------------- THEOREM ---------------
%%-------------------------------------

\begin{theorem}[Proposition 9.8~\cite{lovaz2012large}]\label{thm_partition_conv_book}
	Let $\{ \ccalP_i \}_i$ be a sequence of measureable partitions of $[0,1]$ such that every pair of points is separated by all but a finite number of partitions $\ccalP_i$. Then, $W_{\ccalF_i}\rightarrow W(x,y)$ almost everywhere for every $W\in\ccalW$, where $\ccalW$ is the space of kernels on $[0,1]^2$.
\end{theorem}

%%---------- End of Theorem --------------

 The graphs built from methods \textbf{M1} and \textbf{M2} are steppings of the graphon $W$. Additionally, we remark that every graph obtained using \textbf{M1} and \textbf{M2} is associated to nested partitions of $[0,1]$. This is, if $I_{\ell}^{(i)}$ are the intervals in a partition of $[0,1]$ associated to $\mathcal{P}_{i}$ and $I_{\ell}^{(j)}$ are the intervals in a partition associated to $\mathcal{P}_{j}$, we have that $I_{\ell}^{(i)}=\bigcup_{\ell\in\mathcal{J}}I_{\ell}^{(j)}$, where $\mathcal{J}\subset \mathbb{N}$. Then, the condition stipulated in Theorem~\ref{thm_partition_conv_book} is satisfied and therefore, the sequence of graphons $\{  W_{G_i}  \}_i$ converges almost everywhere to $W$.

Now, from Lemma 9.11 in~\cite{lovaz2012large} we have

	\begin{equation}
	  \left\Vert 
	        W
	        -
	        W_{G_\ell}
	  \right\Vert_{\square}  
	                   \leq 
	                    \frac{2}{\sqrt{\log \vert V(G_\ell ) \vert}}
	                    .
	\end{equation}

Taking into account that $\Vert \boldsymbol{T}_W \Vert_{\infty\rightarrow 1} \leq 4\Vert W\Vert_{\square}$ -- see \cite{lovaz2012large} page 134 -- it follows that

	\begin{equation}
	  \left\Vert 
	        \boldsymbol{T}_{W}
	        -
	        \boldsymbol{T}_{W_{G_\ell}}
	  \right\Vert_{\infty\rightarrow 1}  
	                   \leq 
	                    \frac{8}{\sqrt{\log \vert V(G_\ell ) \vert}}
	                    .
	\end{equation}
\end{proof}

%%%%%%%%%%%%%%%%%%%%%%%%%%%%%%%
%%%%%%%%% Sub-Section %%%%%%%%%
%%%%%%%%%%%%%%%%%%%%%%%%%%%%%%%

\subsection{Fr\'echet derivative of a Graph/Graphon Filter}
\label{sec_Frechet_derivative_GraphGraphon_filter}

In this section we show the details for the calculation of the Fr\'echet derivative of a graph filter instantiated in the induced graphon representation.

%%-------------------------------------
%%------------ THEOREM ----------------
%%-------------------------------------

\begin{theorem}\label{thm_explicitform_frechet_der}

Let $\boldsymbol{T}_W$ be the graphon shift operator of the graphon $W$, and let $h(\boldsymbol{T}_W) = \sum_{k=0}^{\infty}h_k \boldsymbol{T}_{W}^k$. If $D_{h\vert\boldsymbol{T}_W}\{ \boldsymbol{\xi} \}$ is the Fr\'echet derivative of $h(\boldsymbol{T}_W)$, acting on $\boldsymbol{\xi}$, it follows that

\begin{equation}\label{eq_proof_theorem_2}
D_{h\vert\boldsymbol{T}_{W}}\{  \boldsymbol{\xi}  \} 
                          =
                          \sum_{i=1}^{\infty}
                                   \ell_i (\boldsymbol{T}_W) 
                                                         \boldsymbol{\xi}
                                   r_i (\boldsymbol{T}_W) 
                                   ,
\end{equation}
where $\ell_i (\boldsymbol{T}_W)$ and $r_i (\boldsymbol{T}_W)$ are monomial functions.
\end{theorem}

%%--------- End of Theorem ------------

\begin{proof}

Let us start taking into account that

\begin{equation}
 h(\boldsymbol{T}_W + \boldsymbol{\xi}) - h(\boldsymbol{T}_W)
=
\sum_{k=0}^{\infty}h_{k}\left(\boldsymbol{T}_{W} + \boldsymbol{\xi}\right)^{k}
-
\sum_{k=0}^{\infty}h_{k}\boldsymbol{T}_{W}^{k}
.
\end{equation}
We expand the polynomial expression and re-group the monomials in two terms. One term where $\boldsymbol{\xi}$ pears once and another term where $\boldsymbol{\xi}$ appears more than once. Then, we have
\begin{equation}
  h(\boldsymbol{T}_W + \boldsymbol{\xi}) - h(\boldsymbol{T}_W)
=
\sum_{k=0}^{\infty}h_k \sum_{\ell=1}^{k}
           \boldsymbol{T}_{W}^{\ell-1}\boldsymbol{\xi}\boldsymbol{T}_{W}^{k-\ell}
           +
           o\left( 
                 \Vert \boldsymbol{\xi} \Vert_2
            \right)
,  
\end{equation}
with $o\left(\Vert \boldsymbol{\xi} \Vert_2\right)\rightarrow 0$ as $\Vert \boldsymbol{\xi}\Vert_2 \rightarrow 0$. Notice that the term $o\left(\Vert \boldsymbol{\xi} \Vert_2 \right)$ is a polynomial function where $\boldsymbol{\xi}$ appears more than once in each monomial. Since
$
\sum_{k=0}^{\infty}h_k \sum_{\ell=1}^{k}
           \boldsymbol{T}_{W}^{\ell-1}\boldsymbol{\xi}\boldsymbol{T}_{W}^{k-\ell}
$
is linear and bounded as an operator on $\boldsymbol{\xi}$ we have from the definition of the Fr\'echet derivative~\cite{benyamini2000geometric,lindenstrauss2012frechet} that

\begin{equation}
 D_{h\vert\boldsymbol{T}_{W}}\{  \boldsymbol{\xi}  \} 
 =
 \sum_{k=0}^{\infty}h_k \sum_{\ell=1}^{k}
           \boldsymbol{T}_{W}^{\ell-1}\boldsymbol{\xi}\boldsymbol{T}_{W}^{k-\ell}
           ,
\end{equation}
which is indeed a sum of monomials in $\boldsymbol{T}_W$ acting on the left and right of $\boldsymbol{\xi}$.
\end{proof}

%%%%%%%%%%%%%%%%%%%%%%%%%%%%
%%%%%%%%% Sub-Section %%%%%%%%%
%%%%%%%%%%%%%%%%%%%%%%%%%%%%

\subsection{Proof of Theorem~\ref{thm_change_filters}}
\label{sec_proof_thm_change_filters}

Let $\boldsymbol{\xi} = \boldsymbol{T}_{W} - \boldsymbol{T}_{W_G}$ and $D_{h\vert\boldsymbol{T}_{W}}\{  \boldsymbol{\xi}  \} $ be the Fr\'echet derivative of $h: \ccalB (L^2[0,1]) \rightarrow \ccalB (L^2[0,1]) $. Then, from the definition of Fr\'echet derivative -- see proof of Theorem~\ref{thm_explicitform_frechet_der} in Section~\ref{sec_Frechet_derivative_GraphGraphon_filter} -- we have

\begin{equation}\label{eq_proof_theorem_0}
                 h\left(\boldsymbol{T}_W \right)
                 -
                 h\left(\boldsymbol{T}_{W_G}\right)
=
D_{h\vert\boldsymbol{T}_{W}}\{  \boldsymbol{\xi}  \} 
+
o\left( \Vert\boldsymbol{\xi}\Vert_2 \right)
                ,
\end{equation}
where $D_{h\vert\boldsymbol{T}_{W}}\{  \cdot \}$ is linear and bounded. Taking the norm on both sides of the equation above and using the triangle inequality we have

\begin{equation}\label{eq_proof_theorem_1}
\left\Vert 
                 h\left(\boldsymbol{T}_W \right)
                 -
                 h\left(\boldsymbol{T}_{W_G}\right)
\right\Vert_2                 
                 \leq
\left\Vert                  
                 D_{h\vert\boldsymbol{T}_{W}}\{  \boldsymbol{\xi}  \} 
\right\Vert_2               
                 +
                 \ccalO\left( \Vert\boldsymbol{\xi}\Vert_{2}^2 \right)
                 .
\end{equation} 

Now, we focus our attention on the term $ D_{h\vert\boldsymbol{T}_{W}}\{  \boldsymbol{\xi}  \} $. Since $h(\cdot)$ is a polynomial function, from Theorem~\ref{thm_explicitform_frechet_der} we can express $ D_{h\vert\boldsymbol{T}_{W}}\{  \boldsymbol{\xi}  \} $ as 
$
D_{h\vert\boldsymbol{T}_{W}}\{  \boldsymbol{\xi}  \} 
                          =
                          \sum_{i=1}^{\infty}
                                   \ell_i (\boldsymbol{T}_W) 
                                                         \boldsymbol{\xi}
                                   r_i (\boldsymbol{T}_W) 
                                   ,
$
where $\ell_i$ and $r_i$ are polynomial functions. Since $\boldsymbol{T}_W$ is Hilbert-Schmidt, $
D_{h\vert\boldsymbol{T}_{W}}\{  \cdot  \} $ is also Hilbert-Schmidt. Then, as stated in~\cite{conway2019course} (page 268) there is an isomorphic-isometric image of $D_{h\vert\boldsymbol{T}_{W}}\{  \cdot \} \in\ccalB\left( \ccalB \left( L^2 [0,1] \right)\right)$ in $\ccalB (L^{2}[0,1])^{\ast}\otimes\ccalB (L^{2}[0,1])$ that we denote by $\overline{D}_{h\vert\boldsymbol{T}_{W}}\{  \cdot \} $. From~\cite{higham2008functions} (page 61), the eigenvalues of $\overline{D}_{h\vert\boldsymbol{T}_{W}}\{  \cdot \}$ can be written as
\begin{multline}\label{eq_proof_theorem_3}
\lambda_{i,j}\left( 
\overline{D}_{h\vert\boldsymbol{T}_{W}}\{  \cdot \} 
\right)
=
\\
\\
\begin{cases}
           \frac{
           	       h\left( \lambda_i (\boldsymbol{T}_W)\right) - h\left( \lambda_j (\boldsymbol{T}_W)\right)  
           	    }
           	    {
           	    	\lambda_i (\boldsymbol{T}_W) - \lambda_j (\boldsymbol{T}_W)
           	    	}
           	    	,
           	    	\quad
           	    	\lambda_i (\boldsymbol{T}_W) \neq \lambda_j (\boldsymbol{T}_W)
                 \\
                h^{'}\left( \lambda_i (\boldsymbol{T}_W)  \right)
                ,
                \quad 
                \lambda_i (\boldsymbol{T}_W) = \lambda_j (\boldsymbol{T}_W)
                ,
\end{cases}
\end{multline}
while the eigenvectors of $\overline{D}_{h\vert\boldsymbol{T}_{W}}\{  \cdot \} $ are given by $\boldsymbol{\varphi}_{i,j}=\boldsymbol{
	\varphi}_i \otimes \boldsymbol{\varphi}_j$. If $\boldsymbol{\xi}\in\ccalB (L^2[0,1])$ is Hilbert-Schmidt, there exists $\overline{\boldsymbol{\xi}}\in L^{2}[0,1]\otimes L^{2}[0,1]$ defined by the isomorphic-isometric map between $\ccalB \left( \ccalB \left( L^2 [0,1] \right)\right) $ and  $\ccalB (L^{2}[0,1])^{\ast}\otimes\ccalB (L^{2}[0,1])$. 

With these facts at hand we start taking into account that from the operator norm it follows

\begin{equation}\label{eq_proof_theorem_4}
\left\Vert                  
                D_{h\vert\boldsymbol{T}_{W}}\{  \boldsymbol{\xi}  \} 
\right\Vert_2 
                  \leq 
 \left\Vert                  
           D_{h\vert\boldsymbol{T}_{W}}\{  \cdot  \} 
 \right\Vert_{op2}    
  \Vert \boldsymbol{\xi} \Vert_2          
  .    
\end{equation}

Then, since 
$
\left\Vert                  
            D_{h\vert\boldsymbol{T}_{W}}\{  \cdot  \} 
\right\Vert_{op2} 
=
\left\Vert                  
            \overline{D}_{h\vert\boldsymbol{T}_{W}}\{  \cdot  \} 
\right\Vert_{op(\ccalB\otimes\ccalB)} 
,
$
and $\Vert \boldsymbol{\xi} \Vert_2 \leq \Vert \boldsymbol{\xi}\Vert_{HS} = \Vert \overline{\boldsymbol{\xi}}\Vert_{L^2 \otimes L^2}$, we have

\begin{equation}\label{eq_proof_theorem_5}
\left\Vert                  
                D_{h\vert\boldsymbol{T}_{W}}\{  \boldsymbol{\xi}  \} 
\right\Vert_2 
                  \leq 
\left\Vert                  
                \overline{D}_{h\vert\boldsymbol{T}_{W}}\{  \cdot  \} 
\right\Vert_{op(\ccalB\otimes\ccalB)} 
\left\Vert
            \boldsymbol{\xi}
\right\Vert_{HS}
.
\end{equation}

Replacing~\eqref{eq_proof_theorem_5} in~\eqref{eq_proof_theorem_1} we have
\begin{multline}\label{eq_proof_theorem_6}
\left\Vert 
                 h\left(\boldsymbol{T}_W \right)
                 -
                 h\left(\boldsymbol{T}_{W_G}\right)
\right\Vert_2                 
                 \leq
                 \\
\left\Vert                  
                \overline{D}_{h\vert\boldsymbol{T}_{W}}\{  \cdot  \} 
\right\Vert_{op (\ccalB\otimes\ccalB)} 
\left\Vert
            \boldsymbol{\xi}
\right\Vert_{HS}              
                 +
                 \ccalO\left( \Vert\boldsymbol{\xi}\Vert_{2}^2 \right)
                 .
\end{multline} 
Since $h(t)$ is $L$-Lipschitz, from~\eqref{eq_proof_theorem_3} we have that 
$
\left\Vert                  
                \overline{D}_{h\vert\boldsymbol{T}_{W}}\{  \cdot  \} 
\right\Vert_{op (\ccalB \otimes \ccalB)} 
\leq 
L
$
, and therefore \eqref{eq_proof_theorem_6} turns into

\begin{equation}
\left\Vert 
                 h\left(\boldsymbol{T}_W \right)
                 -
                 h\left(\boldsymbol{T}_{W_G}\right)
\right\Vert_2                 
                 \leq
                 L
\left\Vert
            \boldsymbol{\xi}
\right\Vert_{HS}              
                 +
                 \ccalO\left( \Vert\boldsymbol{\xi}\Vert_{2}^2 \right)
                 .
\end{equation} 

Since 
$
\left\Vert
            \boldsymbol{\xi}
\right\Vert_{HS} 
\leq
\gamma
\left\Vert
            \boldsymbol{\xi}
\right\Vert_{2} 
$ 
and
$
\left\Vert 
           \boldsymbol{T}_{W}
\right\Vert_{2}
\leq 
\sqrt{
8\Vert W \Vert_{\square}
}
$
(see Proposition 4, page 15 in~\cite{ruiz2020graphon}) we have

\begin{equation}
\left\Vert 
                 h\left(\boldsymbol{T}_W \right)
                 -
                 h\left(\boldsymbol{T}_{W_G}\right)
\right\Vert_2                 
                 \leq
                 L\gamma
\sqrt{8                 
\left\Vert
            W - W_{G}
\right\Vert_{\square} 
}
                 +
                 \ccalO\left( \Vert\boldsymbol{\xi}\Vert_{2}^2 \right)
                 .
\end{equation} 

Taking into account Lemma~\ref{lemma_cutmetric_zeroerror_step}, there exists a graph $H$ with $V(H)\leq 4 V(G)$ such that

\begin{equation}
\left\Vert 
                 h\left(\boldsymbol{T}_W \right)
                 -
                 h\left(\boldsymbol{T}_{W_G}\right)
\right\Vert_2                 
                 \leq
                 L\gamma
\sqrt{8                 
\left\Vert
            W_{H} - W_{G}
\right\Vert_{\square} 
}
                 +
                 \ccalO\left( \Vert\boldsymbol{\xi}\Vert_{2}^2 \right)
                 .
\end{equation} 
%
%
%

%%%%%%%%%%%%%%%%%%%%%%%%%%%%%%%%%%%%%%%%%%%%%
%%%%%%%%%%%%%% SUB-SECTION %%%%%%%%%%%%%%%%%%
%%%%%%%%%%%%%%%%%%%%%%%%%%%%%%%%%%%%%%%%%%%%%	
	
\subsection{Proof of Theorem~\ref{theorem:stabilityAlgNN0}}
\label{prooftheorem:stabilityAlgNN0}

\begin{proof}

First, we are going to estimate the upper bound of the difference between the perceptron operators in each layer of the graphon neural networks. Since $\eta_\ell$ is $1$-Lipschitz we have

	\begin{multline}
	\left\Vert 
	\eta_{\ell}
	\left(
	h_\ell (\boldsymbol{T}_{W})
	\boldsymbol{x}_{\ell-1}\right)
	-
	\eta_{\ell}
	\left(
	h_\ell (\boldsymbol{T}_{W_{G_\ell}})
	\boldsymbol{x}_{\ell-1}
	\right)
	\right\Vert_2
	\\
	\leq
	\left\Vert 
	h_\ell (\boldsymbol{T}_{W})
	\boldsymbol{x}_{\ell-1}
	-
	h_\ell (\boldsymbol{T}_{W_{G_\ell}})
	\boldsymbol{x}_{\ell-1}
	\right\Vert_2
	.
	\end{multline}
Now, by means of the operator norm property we have

	\begin{multline}
	\left\Vert 
	\eta_{\ell}
	\left(
	h_\ell (\boldsymbol{T}_{W})
	\boldsymbol{x}_{\ell-1}\right)
	-
	\eta_{\ell}
	\left(
	h_\ell (\boldsymbol{T}_{W_{G_\ell}})
	\boldsymbol{x}_{\ell-1}
	\right)
	\right\Vert_2
	\\
	\leq
	\left\Vert 
	h_\ell (\boldsymbol{T}_{W})
	-
	h_\ell (\boldsymbol{T}_{W_{G_\ell}})
	\right\Vert_2
	\Vert
	     \boldsymbol{x}_{\ell-1}
	\Vert_2
	.
	\end{multline}
In what follows we will use the notation 
$
\boldsymbol{E}_{\ell} = 
	\left\Vert 
	h_\ell (\boldsymbol{T}_{W})
	-
	h_\ell (\boldsymbol{T}_{W_{G_\ell}})
	\right\Vert_2
	.
$

Now, we analyze the difference between the output signals in the $\ell$-th layer, which we can write as follows	
	\begin{multline}
	\left\Vert
	\boldsymbol{x}_{\ell}
	-
	\tilde{\boldsymbol{x}}_{\ell}
	\right\Vert_2
	\leq
	\left\Vert
	   \eta_{\ell-1}h_{\ell-1}(\boldsymbol{T}_{W})
	\eta_{\ell-2}
     h_{\ell-2}(\boldsymbol{T}_{W})
	\cdots
	\right.
	\\
	\eta_1
	h_{1}(\boldsymbol{T}_{W})
	\boldsymbol{x}
	-
	\\
	\left. 
	\eta_{\ell-1}h_{\ell-1}(\boldsymbol{T}_{W_{G_{\ell-1}}})
	\eta_{\ell-2}
	h_{\ell-2}(\boldsymbol{T}_{W_{G_{\ell-2}}})\cdots
	\right.
	\\
	\left.
	\eta_{1}
	h_{1}(\boldsymbol{T}_{W_{G_{1}}})
	\boldsymbol{x}
	\right\Vert_2.
	\label{eq:longalgNNexp1}
	\end{multline}

	Then, we take into account that
	\begin{multline}\label{eq:longalgNNexp2}
	h_{\ell+1}(\boldsymbol{T}_W)\eta_{\ell}(a)
	-
	h_{\ell+1}(\boldsymbol{T}_{W_{G_{\ell+1}}})\eta_{\ell}(\tilde{a})
	=
	\\
	(h_{\ell+1}(\boldsymbol{T}_W)
	-
	h_{\ell+1}(\boldsymbol{T}_{W_{G_{\ell+1}}}))\eta_{\ell}(a)
	+
	\\
	h_{\ell+1}(\boldsymbol{T}_{W_{G_\ell}})(\eta_{\ell}(a)-\eta_{\ell}(\tilde{a})),
	\end{multline}
	where $a$ and $\tilde{a}$ indicate the terms that are on the right side of $\eta_{\ell}$ in~\eqref{eq:longalgNNexp2}. Since $\Vert\eta_{\ell}(a)-\eta_{\ell}(b)\Vert_2\leq \Vert a-b\Vert_2$,
	$
	\Vert 
	h_{\ell+1}(\boldsymbol{T}_{W})
	-
	h_{\ell+1}(\boldsymbol{T}_{W_{G_{\ell+1}}})
	\Vert_2
	\leq\boldsymbol{E}_{\ell+1}
	$ 
	,
	and $\Vert h_{\ell+1}\Vert_2 \leq 1$, we have that
	\begin{multline}
	\left\Vert
	h_{\ell+1}(\boldsymbol{T}_W)\eta_{\ell}(a)
	-
	h_{\ell+1}(\boldsymbol{T}_{W_{G_{\ell+1}}})\eta_{\ell}(\tilde{a})
	\right\Vert_2
	\leq
	\\
    \boldsymbol{E}_{\ell+1}
	\Vert a\Vert_2
	+
	\Vert a-\tilde{a}\Vert_2
	.
	\end{multline}

	Taking into account these results recursively on the index $\ell$ we have
	
% 	%
% 	%
% 	%
% 	\begin{equation}\label{eq:aux1}
% 	\left\Vert
% 	\boldsymbol{x}_{L}-\tilde{\boldsymbol{x}}_{L}
% 	\right\Vert_2
% 	\leq
% 	\sum_{\ell=1}^{L}\boldsymbol{E}_{\ell}
% 	\left\Vert
% 	      \boldsymbol{x}
% 	\right\Vert_2
% 	,
% 	\end{equation}
% 	%
% 	%
% 	%
	
%     From ~\eqref{eq:aux1} it follows that
    
% 	%
% 	%
% 	%
% 	\begin{multline}
% 	\left\Vert
% 	   \Phi\left(\boldsymbol{x}, \{ \mathcal{P}_\ell \}_{\ell=1}^{L} , \boldsymbol{T}_{W}\right)
% 	   -
% 	  \Phi\left(\boldsymbol{x},\{ \mathcal{P}_\ell \}_{\ell=1}^{L},\{ \boldsymbol{T}_{W_{G_\ell}}\}_{1}^{L}\right)
% 	\right\Vert_{2}^{2}
% 	\\
% 	=
%     \left\Vert\boldsymbol{x}_{L}-\tilde{\boldsymbol{x}}_{L}
% 	\right\Vert_{2}^{2}
% 	,
% 	\end{multline}
% 	%
% 	%
% 	%
% 	and therefore
	%
	%
	%
	\begin{multline}
	\left\Vert
	\Phi\left(\boldsymbol{x},\{ \mathcal{P}_{\ell} \}_{\ell=1}^{L},\{ \mathcal{S}_{\ell}\}_{\ell=1}^{L}\right)-
	\Phi\left(\boldsymbol{x},\{ \mathcal{P}_{\ell} \}_{\ell=1}^{L},\{ \tilde{\mathcal{S}}_{\ell}\}_{\ell=1}^{L}\right)
	\right\Vert_2
	\\
	\leq
	\sum_{\ell=1}^{L}\boldsymbol{E}_{\ell}
	\left\Vert
	     \boldsymbol{x}
	\right\Vert_2
	.
	\end{multline}

	Finally, we take into account that
	$
\boldsymbol{E}_{\ell} = 
	\left\Vert 
	h_\ell (\boldsymbol{T}_{W})
	-
	h_\ell (\boldsymbol{T}_{W_{G_\ell}})
	\right\Vert_2
$ 
is bounded according to Theorem~\ref{thm_change_filters}. This completes the proof.
\end{proof}

%%%%%%%%%%%%%%%%%%%%%%%%%%%%%%%%%
%%%%%%%%% Sub-Section %%%%%%%%%
%%%%%%%%%%%%%%%%%%%%%%%%%%%%%%%%%

\subsection{Proof of Theorem~\ref{thm_estimate_graphons}}
\label{sec_proof_thm_estimate_graphons}

\begin{proof}

First we take into account that by means of Lemma 5 in~\cite{ruiz2020graphon} (page 16) we have
$
    \Vert W - W_{G_i} \Vert_2
    =
    \Vert 
        \boldsymbol{T}_{W}
        -
        \boldsymbol{T}_{W_{G_i}}
    \Vert_{HS}      
    .
$
Then, taking into account that $\Vert W \Vert_1 \leq \Vert W \Vert_2$ for any graphon $W$ -- see~\cite{lovaz2012large}, page 131 -- it follows that
$
\left\Vert 
       W - W_{G_i}
\right\Vert_1
      \leq 
      \epsilon
      .
$

Now, since $\Vert W \Vert_{\square}\leq \Vert \boldsymbol{T}_W \Vert_{\infty\rightarrow 1}$ -- see~\cite{lovaz2012large}, page 134 -- we have 
\begin{equation}
\left\Vert
    W_{G_i}
    -
    W_{H_i}
\right\Vert_{\square}
\leq 
\frac{\epsilon}{V(H_i)^4}
.
\end{equation}

Then, taking into account Theorem 9.32 in~\cite{lovaz2012large} we have

\begin{equation}
     \left\Vert 
            W_{H_1} - W_{H_2}^{\theta}
     \right\Vert_{1}
     \leq 8\epsilon
     .
\end{equation}

Finally, since $\Vert \boldsymbol{T}_{W} \Vert_{\infty\rightarrow 1}\leq 4\Vert W\Vert_{\square}$ and $\Vert W \Vert_{\square}\leq \Vert W\Vert_1$ we have

\begin{equation}
     \left\Vert 
            \boldsymbol{T}_{W_{H_1}} - \boldsymbol{T}_{W_{H_2}^{\theta}}
     \right\Vert_{\infty\rightarrow 1}
     \leq 32\epsilon
.
\end{equation}

\end{proof}

%%%%%%%%%%%%%%%%%%%%%%%%%%%%%%%%%%%%%%%%%%
%%%%%%%%%%% SUB - SECTION %%%%%%%%%%%%%%%%
%%%%%%%%%%%%%%%%%%%%%%%%%%%%%%%%%%%%%%%%%%

\subsection{Proof of Theorem~\ref{thm_edge_drooping}}
\label{proof_thm_edge_drooping}

First, we start taking into account that
\begin{multline}
\left\Vert 
      h\left( 
          \boldsymbol{T}_{W}
       \right)
      -
      h\left( 
          \widehat{\boldsymbol{T}}_{W_{\widehat{G}}}
       \right)
\right\Vert_{2}
               =
 \left\Vert 
      h\left( 
          \boldsymbol{T}_{W}
       \right)
      -
      h\left( 
          \boldsymbol{T}_{W_{G}}
       \right)
       \right.
       \\
       \left.
             +
             \left(
                   h\left( 
                       \boldsymbol{T}_{W_{G}}
                    \right)
                   -
                   h\left( 
                        \widehat{\boldsymbol{T}}_{W_{\widehat{G}}}
                    \right)
             \right)
\right\Vert_{2} 
.
\end{multline}
Then, taking into account the triangular inequality we have
\begin{multline}
\left\Vert 
      h\left( 
          \boldsymbol{T}_{W}
       \right)
      -
      h\left( 
          \widehat{\boldsymbol{T}}_{W_{\widehat{G}}}
       \right)
\right\Vert_{2}
               \leq
 \left\Vert 
      h\left( 
          \boldsymbol{T}_{W}
       \right)
      -
      h\left( 
          \boldsymbol{T}_{W_{G}}
       \right)
       \right\Vert_{2}
       \\
+
       \left\Vert
                   h\left( 
                       \boldsymbol{T}_{W_{G}}
                    \right)
                   -
                   h\left( 
                        \widehat{\boldsymbol{T}}_{W_{\widehat{G}}}
                    \right)
        \right\Vert_{2} 
.
\end{multline}

If we choose $h(t)=t$, we obtain~\eqref{eq_thm_edge_dropping_1}. On the other hand, if we choose $h(t)$ to be a $C$-Lipschitz filter, by Theorem~\ref{thm_change_filters} we have that
\begin{equation}
\left\Vert 
               h\left(\boldsymbol{T}_W \right)
                 -
               h\left(\boldsymbol{T}_{W_G}\right)
\right\Vert_2                 
               \leq
               \Omega \left( \bbT_{W} \right)
                +
               \ccalO\left( \Vert
                                  \boldsymbol{T}_{W}-\boldsymbol{T}_{W_{G}}
                               \Vert_{2}^2 
                        \right)
                ,
\end{equation} 
and by means of Theorem~1, Theorem~2 and Corollary~1 in~\cite{alejo_algnn_j} we have that
\begin{equation}
       \left\Vert
                   h\left( 
                       \boldsymbol{T}_{W_{G}}
                    \right)
                   -
                   h\left( 
                        \widehat{\boldsymbol{T}}_{W_{\widehat{G}}}
                    \right)
        \right\Vert_{2} 
        \leq 
        (1+\delta)C\Vert 
                       \boldsymbol{T}_{0}
                   \Vert_2   
                   ,
\end{equation}
which completes the proof.

%% file: compiler_paper.bbl
\begin{thebibliography}{10}

\bibitem{alej2020graphon}
Alejandro Parada-Mayorga, Luana Ruiz, and Alejandro Ribeiro,
\newblock ``Graphon pooling in graph neural networks,''
\newblock in {\em 2020 28th European Signal Processing Conference (EUSIPCO)},
  2021, pp. 860--864.

\bibitem{boureau_pooling_1}
Y-Lan Boureau, Jean Ponce, and Yann LeCun,
\newblock ``A theoretical analysis of feature pooling in visual recognition,''
\newblock in {\em Proceedings of the 27th International Conference on
  International Conference on Machine Learning}, Madison, WI, USA, 2010,
  ICML'10, p. 111–118, Omnipress.

\bibitem{boureau_pooling_2}
Y-Lan Boureau, Nicolas Le~Roux, Francis Bach, Jean Ponce, and Yann LeCun,
\newblock ``Ask the locals: Multi-way local pooling for image recognition,''
\newblock in {\em 2011 International Conference on Computer Vision}, 2011, pp.
  2651--2658.

\bibitem{goodfellow2016deep}
I.~Goodfellow, Y.~Bengio, and A.~Courville,
\newblock {\em Deep Learning},
\newblock Adaptive Computation and Machine Learning series. MIT Press, 2016.

\bibitem{mathDNN_wiatowski}
Thomas Wiatowski and Helmut Bölcskei,
\newblock ``A mathematical theory of deep convolutional neural networks for
  feature extraction,''
\newblock {\em IEEE Transactions on Information Theory}, vol. 64, no. 3, pp.
  1845--1866, 2018.

\bibitem{Bronstein2021GeometricDL}
Michael~M. Bronstein, Joan Bruna, Taco Cohen, and Petar Velivckovi'c,
\newblock ``Geometric deep learning: Grids, groups, graphs, geodesics, and
  gauges,''
\newblock {\em ArXiv}, vol. abs/2104.13478, 2021.

\bibitem{oppenheim1975digital}
A.V. Oppenheim and R.W. Schafer,
\newblock {\em Digital Signal Processing},
\newblock MIT video course. Prentice-Hall, 1975.

\bibitem{ortega_proxies}
A.~Anis, A.~Gadde, and A.~Ortega,
\newblock ``Efficient sampling set selection for bandlimited graph signals
  using graph spectral proxies,''
\newblock {\em IEEE Transactions on Signal Processing}, vol. 64, no. 14, pp.
  3775--3789, July 2016.

\bibitem{tsitsverobarbarossa}
M.~{Tsitsvero}, S.~{Barbarossa}, and P.~{Di Lorenzo},
\newblock ``Signals on graphs: Uncertainty principle and sampling,''
\newblock {\em IEEE Transactions on Signal Processing}, vol. 64, no. 18, pp.
  4845--4860, Sep. 2016.

\bibitem{bookgspchapsampling}
P.~Di Lorenzo, S.~Barbarossa, , and P.~Banelli,
\newblock ``Sampling and recovery of graph signals,''
\newblock in {\em Cooperative and Graph Signal Processing}, P.~Djuric and
  C.Richard, Eds. Elsevier, 2018.

\bibitem{7055883}
X.~{Wang}, P.~{Liu}, and Y.~{Gu},
\newblock ``Local-set-based graph signal reconstruction,''
\newblock {\em IEEE Transactions on Signal Processing}, vol. 63, no. 9, pp.
  2432--2444, May 2015.

\bibitem{7581102}
S.~{Chen}, R.~{Varma}, A.~{Singh}, and J.~{Kovačević},
\newblock ``Signal recovery on graphs: Fundamental limits of sampling
  strategies,''
\newblock {\em IEEE Transactions on Signal and Information Processing over
  Networks}, vol. 2, no. 4, pp. 539--554, Dec 2016.

\bibitem{alejopm_bn_j}
Alejandro Parada-Mayorga, Daniel~L. Lau, Jhony~H. Giraldo, and Gonzalo~R. Arce,
\newblock ``Blue-noise sampling on graphs,''
\newblock {\em IEEE Transactions on Signal and Information Processing over
  Networks}, vol. 5, no. 3, pp. 554--569, 2019.

\bibitem{alejopm_bn_dsw}
Alejandro Parada-Mayorga, Daniel~L. Lau, Jhony~H. Giraldo, and Gonzalo~R. Arce,
\newblock ``Sampling of graph signals with blue noise dithering,''
\newblock in {\em 2019 IEEE Data Science Workshop (DSW)}, 2019, pp. 150--154.

\bibitem{alejopm_bn_sampta}
Alejandro Parada-Mayorga, Daniel~L. Lau, Jhony~H. Giraldo, and Gonzalo~R. Arce,
\newblock ``Blue-noise sampling of signals on graphs,''
\newblock in {\em 2019 13th International conference on Sampling Theory and
  Applications (SampTA)}, 2019, pp. 1--5.

\bibitem{alejopm_cographs_dsw}
Dominique Guillot, Alejandro Parada-Mayorga, Sebastian Cioaba, and Gonzalo~R.
  Arce,
\newblock ``Optimal sampling sets in cographs,''
\newblock in {\em 2019 IEEE Data Science Workshop (DSW)}, 2019, pp. 165--169.

\bibitem{alejopm_phd_thesis}
Alejandro Parada-Mayorga,
\newblock {\em Blue Noise and Optimal Sampling on Graphs},
\newblock Ph.D. thesis, 2019.

\bibitem{ortega2022introduction}
A.~Ortega,
\newblock {\em Introduction to Graph Signal Processing},
\newblock Cambridge University Press, 2022.

\bibitem{gsp_sandryhaila}
Aliaksei Sandryhaila and José M.~F. Moura,
\newblock ``Discrete signal processing on graphs,''
\newblock {\em IEEE Transactions on Signal Processing}, vol. 61, no. 7, pp.
  1644--1656, 2013.

\bibitem{Pschel2006AlgebraicSP}
Markus P{\"u}schel and Jos{\'e} M.~F. Moura,
\newblock ``Algebraic signal processing theory,''
\newblock {\em ArXiv}, vol. abs/cs/0612077, 2006.

\bibitem{gsp_sandryhaila_filters}
Aliaksei Sandryhaila and José M.~F. Moura,
\newblock ``Discrete signal processing on graphs: Graph filters,''
\newblock in {\em 2013 IEEE International Conference on Acoustics, Speech and
  Signal Processing}, 2013, pp. 6163--6166.

\bibitem{ortega_gsp}
A.~{Ortega}, P.~{Frossard}, J.~{Kovačević}, J.~M.~F. {Moura}, and
  P.~{Vandergheynst},
\newblock ``Graph signal processing: Overview, challenges, and applications,''
\newblock {\em Proceedings of the IEEE}, vol. 106, no. 5, pp. 808--828, May
  2018.

\bibitem{gamagnns}
F.~{Gama}, A.~G. {Marques}, G.~{Leus}, and A.~{Ribeiro},
\newblock ``Convolutional neural network architectures for signals supported on
  graphs,''
\newblock {\em IEEE Transactions on Signal Processing}, vol. 67, no. 4, pp.
  1034--1049, 2019.

\bibitem{Defferrard2016ConvolutionalNN}
Micha{\"e}l Defferrard, Xavier Bresson, and Pierre Vandergheynst,
\newblock ``Convolutional neural networks on graphs with fast localized
  spectral filtering,''
\newblock in {\em NIPS}, 2016.

\bibitem{graphcoarse1}
I.~S. {Dhillon}, Y.~{Guan}, and B.~{Kulis},
\newblock ``Weighted graph cuts without eigenvectors a multilevel approach,''
\newblock {\em IEEE Transactions on Pattern Analysis and Machine Intelligence},
  vol. 29, no. 11, pp. 1944--1957, 2007.

\bibitem{Diao2016ModelfreeCO}
Peter Diao, Dominique Guillot, Apoorva Khare, and Bala Rajaratnam,
\newblock ``Model-free consistency of graph partitioning,''
\newblock {\em arXiv: Combinatorics}, 2016.

\bibitem{Morency2020GraphonFG}
Matthew~W. Morency and Geert Leus,
\newblock ``Graphon filters: Graph signal processing in the limit,''
\newblock {\em IEEE Transactions on Signal Processing}, vol. 69, pp.
  1740--1754, 2020.

\bibitem{ruiz2020graphon}
Luana Ruiz, Luiz F.~O. Chamon, and Alejandro Ribeiro,
\newblock ``Graphon signal processing,'' 2020.

\bibitem{fern2019stability}
Fernando Gama, Joan Bruna, and Alejandro Ribeiro,
\newblock ``Stability properties of graph neural networks,'' 2019.

\bibitem{lovaz2012large}
L.~Lov{\'a}sz,
\newblock {\em Large Networks and Graph Limits},
\newblock American Mathematical Society colloquium publications. American
  Mathematical Society, 2012.

\bibitem{Glasscock2015WhatIA}
Daniel Glasscock,
\newblock ``What is a graphon,''
\newblock {\em arXiv: Combinatorics}, 2015.

\bibitem{alejo_algnn_j}
Alejandro Parada-Mayorga and Alejandro Ribeiro,
\newblock ``Algebraic neural networks: Stability to deformations,''
\newblock {\em IEEE Transactions on Signal Processing}, vol. 69, pp.
  3351--3366, 2021.

\bibitem{alejo_algnn_c}
Alejandro Parada-Mayorga and Alejandro Ribeiro,
\newblock ``Stability of algebraic neural networks to small perturbations,''
\newblock in {\em ICASSP 2021 - 2021 IEEE International Conference on
  Acoustics, Speech and Signal Processing (ICASSP)}, 2021, pp. 5205--5209.

\bibitem{Ruiz2020GraphonNN}
Luana Ruiz, Luiz F.~O. Chamon, and Alejandro Ribeiro,
\newblock ``Graphon neural networks and the transferability of graph neural
  networks,''
\newblock {\em ArXiv}, vol. abs/2006.03548, 2020.

\bibitem{kingma2014adam}
Diederik~P Kingma and Jimmy Ba,
\newblock ``Adam: A method for stochastic optimization,''
\newblock {\em arXiv preprint arXiv:1412.6980}, 2014.

\bibitem{xu2021learning}
Hongteng Xu, Dixin Luo, Lawrence Carin, and Hongyuan Zha,
\newblock ``Learning graphons via structured gromov-wasserstein barycenters,''
\newblock in {\em Proceedings of the AAAI Conference on Artificial
  Intelligence}, 2021, vol.~35, pp. 10505--10513.

\bibitem{wu20153d}
Zhirong Wu, Shuran Song, Aditya Khosla, Fisher Yu, Linguang Zhang, Xiaoou Tang,
  and Jianxiong Xiao,
\newblock ``3d shapenets: A deep representation for volumetric shapes,''
\newblock in {\em Proceedings of the IEEE conference on computer vision and
  pattern recognition}, 2015, pp. 1912--1920.

\bibitem{ying2018hierarchical}
Zhitao Ying, Jiaxuan You, Christopher Morris, Xiang Ren, Will Hamilton, and
  Jure Leskovec,
\newblock ``Hierarchical graph representation learning with differentiable
  pooling,''
\newblock {\em Advances in neural information processing systems}, vol. 31,
  2018.

\bibitem{harper2015movielens}
F~Maxwell Harper and Joseph~A Konstan,
\newblock ``The movielens datasets: History and context,''
\newblock {\em Acm transactions on interactive intelligent systems (tiis)},
  vol. 5, no. 4, pp. 1--19, 2015.

\bibitem{huang2018rating}
Weiyu Huang, Antonio~G. Marques, and Alejandro~R. Ribeiro,
\newblock ``Rating prediction via graph signal processing,''
\newblock {\em IEEE Transactions on Signal Processing}, vol. 66, no. 19, pp.
  5066--5081, 2018.

\bibitem{gao2019graphon}
S.~Gao,
\newblock {\em Graphon Control Theory for Linear Systems on Complex Networks
  and Related Topics},
\newblock McGill theses. McGill University Libraries, 2019.

\bibitem{benyamini2000geometric}
Y.~Benyamini and J.~Lindenstrauss,
\newblock {\em Geometric Nonlinear Functional Analysis},
\newblock Number v. 48, no. 1 in American Mathematical Society colloquium
  publications. American Mathematical Society, 2000.

\bibitem{lindenstrauss2012frechet}
J.~Lindenstrauss, D.~Preiss, and J.~Ti{\v{s}}er,
\newblock {\em Frechet Differentiability of Lipschitz Functions and Porous Sets
  in Banach Spaces},
\newblock Annals of Mathematics Studies. Princeton University Press, 2012.

\bibitem{conway2019course}
J.B. Conway,
\newblock {\em A Course in Functional Analysis},
\newblock Graduate Texts in Mathematics. Springer New York, 2019.

\bibitem{higham2008functions}
N.J. Higham,
\newblock {\em Functions of Matrices: Theory and Computation},
\newblock Other Titles in Applied Mathematics. Society for Industrial and
  Applied Mathematics (SIAM, 3600 Market Street, Floor 6, Philadelphia, PA
  19104), 2008.

\end{thebibliography}
